\documentclass[acmtog, authorversion]{acmart}

\usepackage{booktabs} 
\usepackage{epsfig}
\usepackage{graphicx}
\usepackage{amsmath}
\usepackage{multirow}
\usepackage{comment}
\usepackage{enumitem}
\usepackage{subfigure}
\usepackage{wrapfig}

\newcounter{mycounter}

 \newcommand{\src}{ \text{src}}
 \newcommand{\trg}{ \text{trg}}

 \newcommand{\rev}[1]{{#1}} 
 \newcommand{\revrev}[1]{{#1}} 

\newcommand{\bc}{\mathbf{c}}

\newcommand{\bh}{\mathbf{h}}

\newcommand{\bp}{\mathbf{p}}
\newcommand{\bq}{\mathbf{q}}

\newcommand{\bs}{\mathbf{s}}

\newcommand{\bw}{\mathbf{w}}

\newcommand{\by}{\mathbf{y}}

\newcommand{\bA}{\mathbf{A}}

\newcommand{\bF}{\mathbf{F}}

\newcommand{\bQ}{\mathbf{Q}}

\newcommand{\bY}{\mathbf{Y}}

\newcommand{\mL}{\mathcal{L}}

\newcommand{\mR}{\mathcal{R}}

\newcommand{\mN}{\mathcal{N}}

\acmJournal{TOG}
\acmYear{2020}
\acmVolume{39}
\acmNumber{6}
\acmArticle{221}
\acmMonth{12}
\acmDOI{10.1145/3414685.3417774}

\setcopyright{acmcopyright}
\copyrightyear{2020}



\begin{document}

\title{MakeItTalk: Speaker-Aware Talking-Head Animation}

\author{Yang Zhou}
\email{yangzhou@cs.umass.edu}
\affiliation{%
  \institution{University of Massachusetts Amherst}
}

\author{Xintong Han}
\email{hixintonghan@gmail.com}
\affiliation{%
  \institution{Huya Inc}
}

\author{Eli Shechtman}
\email{elishe@adobe.com}
\affiliation{%
  \institution{Adobe Research}
}

\author{Jose Echevarria}
\email{echevarr@adobe.com}
\affiliation{%
  \institution{Adobe Research}
}

\author{Evangelos Kalogerakis}
\email{kalo@cs.umass.edu}
\affiliation{%
  \institution{University of Massachusetts Amherst}
}

\author{Dingzeyu Li}
\email{dinli@adobe.com}
\affiliation{%
  \institution{Adobe Research}
}

\renewcommand{\shortauthors}{Zhou et al.}

\begin{abstract}
We present a method that generates expressive talking-head videos from a single facial image with audio as the only input. In contrast to previous attempts to learn direct mappings from audio to raw pixels for creating talking faces, our method first disentangles the content and speaker information in the input audio signal. The audio content robustly controls the motion of lips and nearby facial regions, while the speaker information determines the specifics of facial expressions and the rest of the talking-head dynamics. Another key component of our method is the prediction of facial landmarks reflecting the speaker-aware dynamics. Based on this intermediate representation, our method works with many portrait images in a single unified framework, including artistic paintings, sketches, 2D cartoon characters,  Japanese mangas, and stylized caricatures.
In addition, our method generalizes well for faces and characters that were not observed during training. We present extensive quantitative and qualitative evaluation of our method, in addition to user studies, demonstrating generated talking-heads of significantly higher quality compared to prior state-of-the-art methods.
\footnote{
Our project page with source code, datasets, and supplementary video is available at
\textcolor{blue}{https://people.umass.edu/yangzhou/MakeItTalk/  }
}
\end{abstract}

\begin{teaserfigure}
\centering
\includegraphics[ width=1\textwidth]{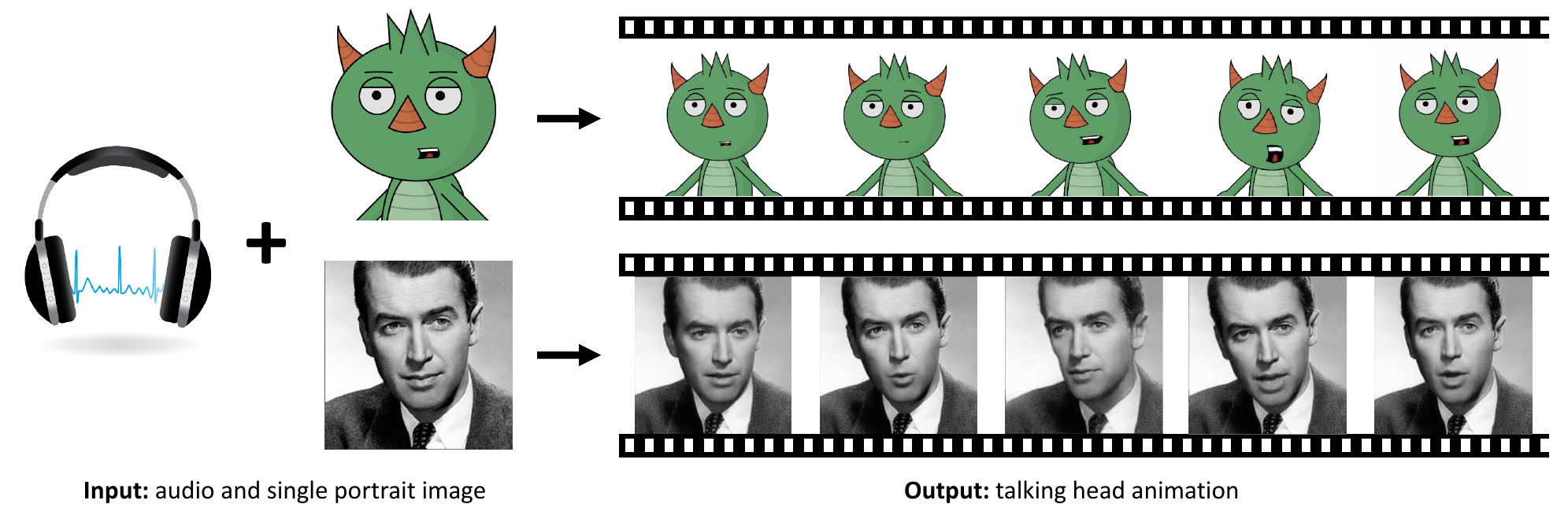}
\vspace{-5mm}
\caption{Given an audio speech signal and a single portrait image   as input (left), our model generates speaker-aware talking-head animations (right). 
\emph{Both the speech signal and the input face image are not observed during the model training process.}
\rev{Our method creates both non-photorealistic cartoon animations (top) and natural human face videos (bottom). Please also see our supplementary video. Cartoon character \emph{Wilk} \textcopyright \emph{Dave Werner} at Adobe Research}. Natural face \emph{James Stewart} by \emph{studio publicity still} (public domain).}
\label{fig:teaser}
\label{fig:gallery}
\end{teaserfigure}

\maketitle

\section{Introduction}
\begin{table*}[t]
\caption{A comparison of related works across to various criteria shown on the left.
``Handle unseen faces'' means handling faces (or rigged models and 3d meshes) unobserved during training. \rev{``Single target image'' means requiring only a single target image instead of a video for talking head generation.}}
\vspace{-3mm}
\small
\centering
\begin{tabular}{c|p{1.6cm}|p{1.4cm}|p{1.5cm}|p{1.4cm}|p{1.2cm}|p{1.4cm}|p{1.2cm}|p{1.2cm}|p{1cm}}
                           & \citet{suwajanakorn2017synthesizing} & \citet{taylor2017deep}    & \citet{karras2017audio}  & \citet{zhou2018visemenet}      & \citet{zhou2019talking} & \citet{vougioukas2019realistic}  & \citet{chen2019hierarchical}  & \citet{thies2019neural}         &    ~Ours                       \\ \hline
animation format           & \hfil image &rigged~model& \hfil 3d mesh &rigged~model &\hfil image  &\hfil image &\hfil image &\hfil image & \multicolumn{1}{c}{image} \\ \cline{2-10} 
beyond lip animation             &\hfil \checkmark     &\hfil $\times$         &\hfil \checkmark       &\hfil $\times$         &\hfil \checkmark      &\hfil \checkmark     &\hfil \checkmark     &\hfil \checkmark             & \multicolumn{1}{c}{\checkmark}     \\ \cline{2-10} 
head pose               &\hfil \checkmark     &\hfil $\times$        &\hfil $\times$      &\hfil $\times$        &\hfil $\times$     &\hfil $\times$    &\hfil $\times$    &\hfil $\times$            & \multicolumn{1}{c}{\checkmark}     \\ \cline{2-10} 
speaker-awareness     &\hfil $\times$    &\hfil $\times$        &\hfil $\times$      &\hfil $\times$        &\hfil $\times$      &\hfil $\times$    &\hfil $\times$    &\hfil \checkmark             & \multicolumn{1}{c}{\checkmark}     \\ \cline{2-10} 
handle unseen faces &\hfil $\times$    &\hfil \checkmark         &\hfil $\times$      &\hfil $\times$        &\hfil \checkmark      &\hfil $\times$    &\hfil \checkmark     &\hfil \checkmark             & \multicolumn{1}{c}{\checkmark}     \\  \cline{2-10} 
\rev{single target image} &\hfil $\times$    &\hfil $-$         &\hfil $-$      &\hfil $-$        &\hfil \checkmark      &\hfil \checkmark    &\hfil \checkmark     &\hfil $\times$             & \multicolumn{1}{c}{\checkmark}     \\  \cline{1-10} 
\end{tabular}
\label{tab:feature_comparisons}
\end{table*}

Animating expressive talking-heads is essential for filmmaking, virtual avatars, video streaming, computer games, and mixed realities. 
Despite recent advances, generating realistic facial animation with little or no manual labor remains an open challenge in computer graphics. 
Several key factors contribute to this challenge.
Traditionally, the \emph{synchronization} between speech and facial movement is hard to achieve manually. 
Facial dynamics lie on a high-dimensional manifold, making it nontrivial to find a mapping from audio/speech \cite{edwards2016jali}.
Secondly, different talking {\emph{styles}} in multiple talking-heads can convey different personalities and lead to better viewing experiences~\cite{walker1997improvising}. 
Last but not least, handling lip syncing and facial animation are not sufficient for the perception of \emph{realism} of talking-heads. The entire facial expression considering the correlation between all facial elements and head pose also play an important role~\cite{faigin2012artist, greenwood2018joint}. 
These correlations, however, are less constrained by the audio and thus hard to be estimated.

In this paper, we propose a new method based on a deep neural architecture, called \emph{MakeItTalk}, to address the above challenges. 
Our method generates talking-heads from a single facial image and audio as the only input. 
At test time, MakeItTalk is able to produce plausible talking-head animations with both facial expressions and head motions for new faces and voices not observed during training.

Mapping audio to facial animation is challenging, since it is not a one-to-one mapping. 
Different speakers can have large variations in head pose and expressions given the same audio content. 
The key insight of our approach 
is to disentangle the speech content and speaker identity information in the input audio signal.
The content captures the phonetic and prosodic information in the input audio and is used for robust \emph{synchronization} of lips and nearby facial regions.
The speaker information captures the rest of the facial expressions and \emph{head motion  dynamics}, \rev{which tend to be characteristic for the speaker and are important for generating expressive talking-head animation.}
We demonstrate that this disentanglement leads to significantly more plausible and believable head animations.
\rev{To deal with the additional challenge of producing coherent head motion, we propose a combination of LSTM and self-attention mechanisms to capture both short and long-range temporal dependencies in head pose.}

Another key component of our method is the prediction of facial landmarks as an intermediate representation incorporating speaker-aware dynamics. 
This is in contrast to previous approaches that attempt to directly generate raw pixels \rev{or 3D morphable face models} from audio. 
Leveraging facial landmarks as the intermediate representation between audio to visual animation has several advantages. 
First, based on our disentangled representations, our model learns to generate landmarks that capture subtle, speaker-dependent dynamics, sidestepping the learning of low-level pixel appearance that tends to miss those.
The degrees of freedom (DoFs) for landmarks is in the order of tens ($68$ in our implementation), as opposed to millions of pixels in raw video generation methods.
As a result, our learned model is also compact, making it possible to train it from moderately sized datasets.
Last but not the least, the landmarks can be easily used to drive a wide range of different types of animation content, including human face images and non-photorealistic cartoon images, such as sketches, 2D cartoon characters, Japanese mangas and stylized caricatures. \rev{In contrast, existing video synthesis methods and morphable models are limited to human faces and cannot readily generalize to non-photorealistic or non-human faces and expressions.}

\rev{Our experiments demonstrate that our method achieves significantly  more accurate and plausible talking heads compared to prior work qualitatively and quantitatively, especially in the challenging setting of animating static face images unseen during training. In addition, our ablation study demonstrates the advantages of disentangling speech content and speaker identity for speaker-aware facial animation.}

In summary, given an audio signal and a single portrait image as input (both unseen during training), our method generates expressive talking-head animations.
We highlight the following contributions:
\begin{itemize}
    \item We introduce a new deep-learning based architecture to predict facial landmarks, capturing both facial expressions and overall head poses, from only speech signals. 
    \item We generate speaker-aware talking-head animations  based on disentangled speech content and speaker information, inspired by advances from voice conversion.
    \item \rev{We present two landmark-driven image synthesis methods for non-photorealistic cartoon images and human face images}. These methods can handle new faces and cartoon characters not observed during training.
    \item We propose a set of quantitative metrics and conduct user studies for  evaluation of talking-head animation methods.
\end{itemize}

\section{Related Work}
\begin{figure*}[t]
\begin{center}
\includegraphics[width=\linewidth]{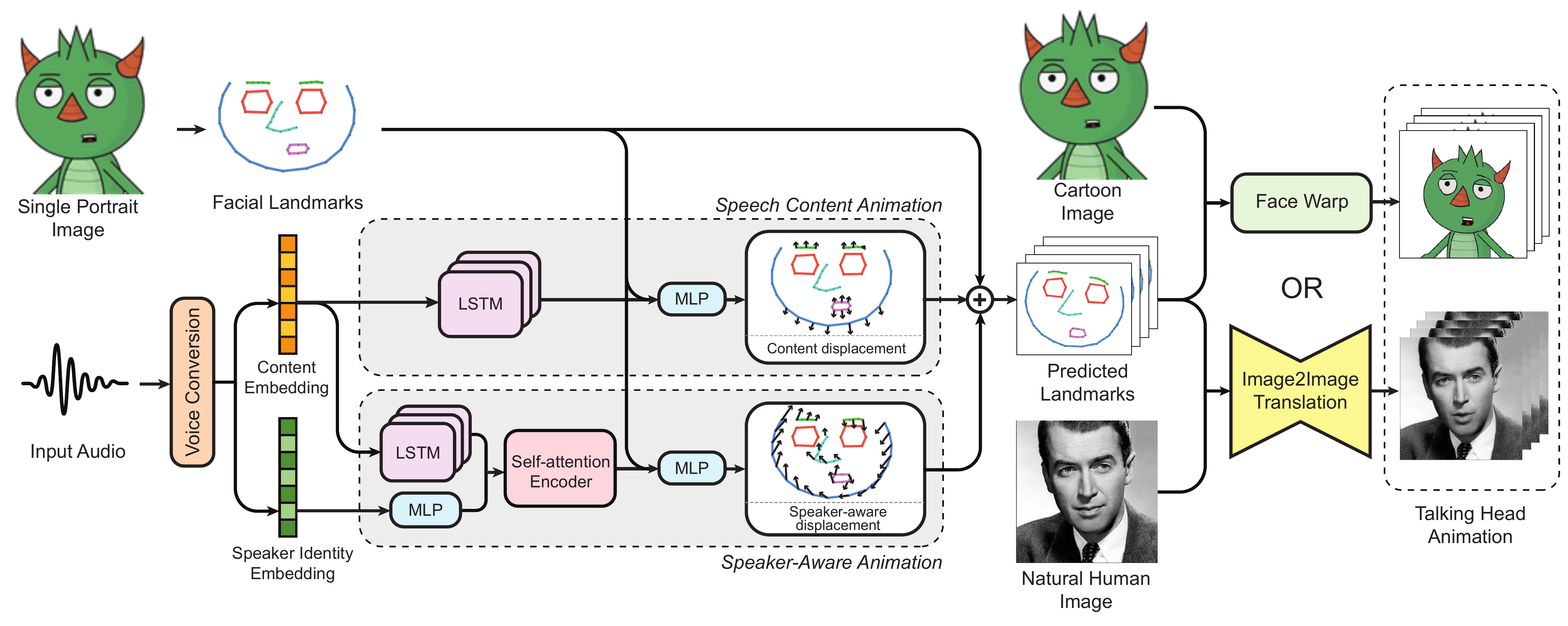}
\end{center}
\vspace{-5mm}
   \caption{Pipeline of our method (``MakeItTalk''). Given an input audio signal along with a single portrait image (cartoon or real photo), our method animates the portrait in a speaker-aware fashion driven by disentangled content and speaker embeddings. 
   The animation is driven by intermediate predictions of 3D landmark displacements. The ``speech content animation'' module maps the disentangled audio content to landmark displacements synchronizing the lip, jaw, and nearby face regions with the input speech. The same set of landmarks is further modulated by the ``speaker-aware animation'' branch that takes into account the speaker embedding to capture the rest of the facial expressions and head motion dynamics.
   }
\label{fig:overview}
\end{figure*}

In computer graphics, we have a long history of cross-modal synthesis. 
More than two decades ago, \citet{brand1999voice} pioneered \emph{Voice Puppetry} to generating full facial animation from an audio track.
Music-driven dance generation from \citet{shiratori2006dancing} matched the rhythm and intensity from the input melody to full-body dance motions. \citet{gan2020foley, gan2020music} separate sound sources and generate synchronized musics from videos with people playing instruments.
Recently, \citet{ginosar2019learning} predicted conversational gestures and skeleton movements from speech signals. 
We focus on audio-driven facial animation, which can supplement other body and gesture prediction methods. In the following paragraphs, we overview prior work on facial landmark synthesis, facial animation, and video generation. Table \ref{tab:feature_comparisons} summarizes differences of methods that are most related to ours based on a set of key criteria.

\paragraph{Audio-driven facial landmark synthesis} 
\citet{eskimez2018generating,eskimez2019noise} 
generated synchronized facial landmarks with robust noise resilience using deep neural networks.
Later, \citet{chen2019hierarchical} trained decoupled blocks to obtain landmarks first and then generate rasterized videos. Attention masks are used to focus on the most changing parts on the face, especially the lips. 
\citet{greenwood2018joint} jointly learnt facial expressions and head poses in terms of landmarks from a forked Bi-directional LSTM network.
Most previous audio-to-face-animation work focused on matching speech content and left out style/identity information since the identity is usually bypassed due to mode collapse or averaging
during training.
In contrast, our approach disentangles audio content and speaker information, and drives landmarks capturing speaker-dependent dynamics.

\paragraph{Lip-sync facial animation} With the increasing power of GPUs, we have seen prolific progress on end-to-end learning from audio to video frames.
\cite{chen2018lip} synthesized cropped lip movements for each frames. \citet{vougioukas2019realistic,song2019,chung2017you} generated full natural human face images with GANs or encoder-decoder CNNs. \cite{pham2018end} estimated blendshape parameters.
\rev{\citet{taylor2017deep, zhou2018visemenet} demonstrated audio-driven talking portraits for rigged face models, however, the input cartoon models required manual rigging and retargeting, as well as artist interventions for  animating the rest of the head beyond lips. Our method is instead able to automatically animate an input portrait and does not require such manual inputs.}
In addition, the above methods do not  capture speaker identity or style. As a result, if the same sentence is spoken by two different voices, they will tend to generate the same facial animation lacking the dynamics required to make it more expressive and realistic.

\paragraph{``Style''-aware facial head animation} \citet{suwajanakorn2017synthesizing} used a re-timing dynamic programming method to reproduce  speaker motion dynamics. However, it was specific to a single  subject (Obama), and does not generalize to faces other than Obama's. \rev{In another earlier work, ~\citet{liu2015video} used color, depth and audio to reproduce the facial animation of a speaker recorded with a RGBD sensor. However, it does not generalize to other unseen speakers.}
\citet{cudeiro2019capture} attempted to model speaker style in a latent representation. 
\citet{thies2019neural} encoded personal style in static blendshape bases.  
Both methods, however, focus on lower facial animation, especially lips, and do not predict head pose.
More similar to ours, \citet{zhou2019talking} learned a joint audio-visual representation to disentangle the identity and content from the image domain.  
However, their identity information primarily focus on static facial appearance and not the speaker dynamics.
As we demonstrate in \S\ref{sec:eval_style}, speaker awareness encompasses many aspects beyond mere static appearances. 
The individual facial expressions and head movements are both important factors for speaker-aware animations.
Our method addresses speaker identity by learning jointly the static appearance and head motion dynamics, to deliver faithfully animated talking-heads. 

\rev{\paragraph{Warpping-based character animation.} \citet{Fiser17npr} and \citet{averbuch2017bringing} demonstrated animation of portraits driven by 
videos and extracted landmarks of human facial performance. \citet{weng2019photo} presented a system for animating the body of an input portrait by fitting a human template, then animated it using motion capture data.
In our case, we aim to synthesize facial expressions and head pose from audio alone. }

\paragraph{Evaluation metrics}
Quantitatively evaluating the learned identity/style is crucial for validation; 
at the same time, it is nontrivial to setup an appropriate benchmark.
Many prior work resorted to subjective user studies~\cite{karras2017audio,cudeiro2019capture}.
\citet{agarwal2019protecting} visualized the style distribution via action units.
For existing quantitative metrics, they primarily focus on pixel-level artifacts since a majority of the network capacity is used to learn pixel generation rather than high-level dynamics~\cite{chen2019hierarchical}. Action units have been proposed to be an alternative evaluation measure of expression in the context of GAN-based approaches~\cite{pumarola2018ganimation}.
We propose a new collection of metrics to evaluate the high-level dynamics that matter for facial expression and head motion. 

\paragraph{Image-to-image translation}
Image-to-image translation is a very common approach to modern talking face synthesis and editing. 
Face2Face and VDub are among the early explorers to demonstrate robust appearance transfer between two talking-head videos \cite{thies2016face2face,garrido2015vdub}. 
Later, adversarial training was adopted to improve the quality of the transferred results.
For example, \citet{kim2019neural} used cycle-consistency loss to transfer styles and showed promising results on one-to-one transfers. \citet{zakharov2019few} developed a few-shot learning scheme that leveraged landmarks to generate natural human facial animation. Based on these prior works, we also employ an image-to-image translation network to generate natural human talking-head animations. Unlike \citet{zakharov2019few}, our model handles generalization to faces unseen during training without the need of fine-tuning. 
Additionally, we are able to generate non-photorealistic images through an image deformation module.

\paragraph{Disentangled learning}
Disentanglement of content and style in audio has been widely studied in the voice conversion community.
Without diving into its long history (see \cite{stylianou2009voice} for a detailed survey), here we only discuss recent methods that fit into our deep learning pipeline.
\citet{wan2018resemblyzer} developed \emph{Resemblyzer} as a speaker identity embedding for verification purposes across  different languages.
\citet{qian2019autovc} proposed AutoVC, a few-shot voice conversion method to separate the audio into the speech content and the identity information. 
As a baseline, we use AutoVC for extracting voice content and Resemblyzer for extracting feature embeddings of  speaker identities.
We introduce the idea of voice conversion to audio-driven animation and demonstrate the advantages of \emph{speaker-aware} talking-head generation.

\section{Method}
\label{sec:method}

\paragraph{Overview} As summarized in Figure~\ref{fig:overview}, given an audio clip and a single facial image, our architecture, called ``MakeItTalk'', generates a \emph{speaker-aware} talking-head animation synchronized with the audio. 
In the training phase, we use an off-the-shelf face 3D landmark detector to preprocess the input videos to extract the landmarks~\cite{bulat2017far}.
A baseline model to animate the speech content can be trained from the input audio and the extracted landmarks directly. However, to achieve high-fidelity dynamics,
we found that landmarks should instead be predicted from a disentangled content representation and speaker embedding of the input audio signal.

Specifically, we use a voice conversion neural network to disentangle the speech content and identity information \cite{qian2019autovc}.
The {\em content} is speaker-agnostic and captures the general motion of lips and nearby regions (Figure~\ref{fig:overview}, \emph{Speech Content Animation}, \S\ref{sec:content_animation}).
The {\em identity} of the speaker determines the specifics of the motions and the rest of the talking-head dynamics (Figure~\ref{fig:overview}, \emph{Speaker-Aware Animation}, \S\ref{sec:style_animation}).
For example, no matter who speaks the word `Ha!', the lips are expected to be open, which is speaker-agnostic and only dictated by the content. 
As for the exact shape and size of the opening, as well as the motion of nose, eyes and head, these will depend on who speaks the word, i.e., identity. Conditioned on the content and speaker identity information, our deep model outputs a sequence of predicted landmarks for the given audio. 

\rev{To generate rasterized images, we discuss two algorithms for the landmark-to-image synthesis (\S\ref{sec:animation}).
For non-photorealistic images like paintings or vector arts (Fig.~\ref{fig:cartoon_examples}), we use a simple image warping method based on Delaunay triangulation inspired by \cite{averbuch2017bringing} (Figure~\ref{fig:overview}, \emph{Face Warp}).
For natural images (Fig.~\ref{fig:natural_examples}), we devised an image-to-image translation network, inspired by pix2pix \cite{isola2017image}) to animate the given human face image with the underlying landmark predictions (Figure~\ref{fig:overview}, \emph{Image2Image Translation}).}
Combining all the image frames and input audio together gives us the final talking-head animations. In the following sections, we describe each module of our architecture.

\subsection{Speech Content Animation}
\label{sec:content_animation}

To extract the speaker-agnostic content representation of the audio, we use AutoVC encoder from
 \citet{qian2019autovc}.
The AutoVC network utilizes an LSTM-based\ encoder that compresses the input audio into a compact  representation (bottleneck) trained to abandon the original speaker identity but preserve content.
In our case, we extract a content embedding $\bA \in \mR^{T \times D}$ from AutoVC network,
where $T$ is the total number of input audio frames, and $D$ is the content dimension. 

The goal of the content animation component is to map the content embedding $\bA$ to facial landmark positions with a neutral style. 
In our experiments, we found that recurrent networks are much better suited for the task than feedforward networks, since they are designed to capture such sequential dependencies between the audio content and landmarks.
We experimented with vanilla RNNs and LSTMs~\cite{graves2014towards}, and found that LSTMs
offered better performance. 
Specifically, at each frame $t$,  the LSTM module takes as input the audio content $\bA$ within a window $[ t \rightarrow t+\tau]$. 
We set $\tau = 18$ frames (a window size of $0.3$s in our experiments). 
To animate any input 3D static landmarks $\bq$, where $\bq \in \mR^{68\times3}$ that are extracted using a landmark detector, the output from LSTM layers is fed into a Multi-Layer Perceptron (MLP) and finally predicts displacements $ \Delta \bq_t$, which put the input landmarks in motion at each frame. 

To summarize, the speech content animation module models sequential dependencies to output landmarks based on
the following transformations: 
\begin{align}
 \bc_t &= LSTM_c \big( \bA_{t \rightarrow t+\tau}; \bw_{lstm,c} \big),  \\
\Delta\bq_t &= MLP_c(\bc_t, \bq;                  \bw_{mlp,c}),  \\
 \bp_t &= \bq + \Delta\bq_t ,
\label{eq:content}
\end{align}
where $\{\bw_{lstm,c}, \bw_{mlp,c}\}$ are learnable parameters for the LSTM and MLP networks respectively. 
The LSTM has three layers of units, each having an internal hidden state vector of size  $256$. 
The decoder MLP network has three layers with internal hidden state vector size of $512$, $256$ and $204$ ($68\times3$), respectively.
 
\subsection{Speaker-Aware Animation}
\label{sec:style_animation}

\begin{figure}[tb]
\begin{center}
\includegraphics[width=1\linewidth]{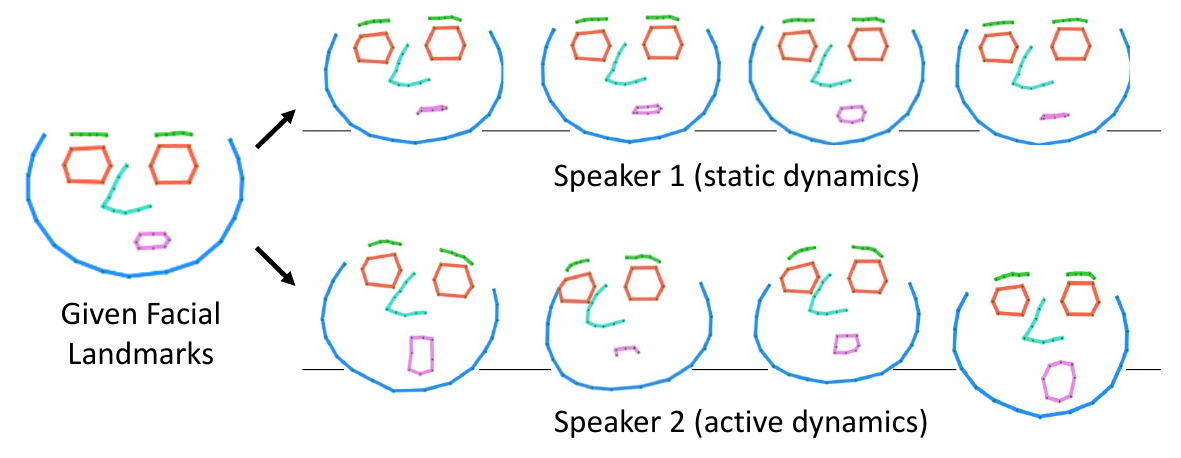}
\end{center}
\vspace{-3mm}
   \caption{Landmark prediction for different speaker identities. Left: static facial landmarks from a given portrait image. Right-top: predicted landmark sequence from a speaker who tends to be conservative in terms of head motion. Right-bottom: predicted landmark sequence from another speaker who tends to be more active.}
\label{fig:trevor_vs_obama}
\end{figure}

Matching just the lip motion to the audio content is not sufficient. 
The motion of the head or the subtle correlation between mouth and eyebrows are also crucial clues to generate plausible talking-heads.
\rev{For example, Figure~\ref{fig:trevor_vs_obama}  shows our speaker-aware predictions for two different speaker embeddings: one originates from a speaker whose head motion tends to be more static, and another that is more active (see also our supplementary video)}. Our method successfully differentiates the head motion dynamics between these two speakers.  

To achieve this, we extract the speaker identity embedding with a speaker verification model~\cite{wan2018resemblyzer} which maximizes the embedding similarity among different utterances of the same speaker, and minimizes the similarity among different speakers. 
The original identity embedding vector $\bs$ has a size of 256. 
We found that  reducing its dimensionality from $256$ to $128$ via a single-layer MLP improved the generalization of facial animations especially for speakers not observed during training. 
Given the identity embedding $\bs$ extracted, we further modulate the per-frame landmarks $\bp_t$ such that they reflect the speaker's identity. 
\rev{More specifically, the landmarks are perturbed to match the head motion distribution and facial expression dynamics observed in speakers during training. In this manner, our method reproduces a  speaker-specific distribution of plausible head movements reflected by the modulated landmarks.}

As shown in the bottom stream of Figure \ref{fig:overview}, we first use an LSTM to encode the content representation within time windows, which has the same network architecture and time window length as the LSTM used in the speech content animation module. We found, however, that it is better to have different learned parameters for this LSTM, such that the resulting representation $\tilde \bc_t$ is more tailored for capturing head motion and facial expression dynamics:
\begin{align}
 \tilde \bc_t &= LSTM_s \big( \bA_{t \rightarrow t+\tau}; \bw_{lstm,s} \big),
\end{align}
where $\bw_{lstm,s}$ are trainable parameters. Then, the following model takes as input 
the speaker embedding  $\bs$, 
the audio content representation $\tilde \bc_t$, 
and the initial static landmarks $\bq$ to generate speaker-aware landmark displacement. 
Notably, we found that producing coherent head motions and facial expressions requires capturing longer time-dependencies compared to the speech content animation module. 
While phonemes typically last for a few tens of milliseconds, head motions, e.g., a head swinging left-to-right, may last for one or few seconds, several magnitudes longer. 
To capture such long and structured dependencies, we adopted a self-attention network~\cite{vaswani2017attention, devlin2018bert}. 
The self-attention layers compute an output expressed as a weighted combination of learned per-frame representations i.e., the audio content representation
$\tilde \bc_t$
extracted by the above LSTM 
concatenated with the speaker embedding $\bs$.
The weight assigned to each frame is computed by a compatibility function  comparing all-pairs frame representations within a window.
We set the window size to $\tau'=256$ frames ($4$ sec) in all experiments. The  output  from the last self-attention layer and the initial static landmarks 
are processed by an MLP to predict the final per-frame landmarks.
 
Mathematically, our speaker-aware animation models structural dependencies to perturb landmarks that capture head motion and personalized expressions, which can be formulated as follows:
\begin{align}
\bh_t& = Attn_s(\tilde \bc_{t \rightarrow t+\tau'}, \bs;       \bw_{\text{attn},s}),   \\
\Delta\bp_t &= MLP_s(\bh_t, \bq;                  \bw_{mlp,s}), \\
 \by_t &= \bp_t + \Delta\bp_t,
\end{align}
where $\{\bw_{\text{attn},s}, \bw_{mlp,c}\}$ are trainable parameters of the self-attention encoder and MLP decoder, $\bp_t$ is computed by Eq. (\ref{eq:content}), and $\by_t$ are the final per-frame landmarks capturing both speech content and speaker identity. 
In our implementation, the attention network follows the encoder block in~\citet{vaswani2017attention}. More details about its architecture are provided in the appendix.

\subsection{Single-Image Animation}
\label{sec:animation}

The last step of our model creates the final animation of the input portrait. 
Given an input image $\bQ$ and the predicted landmarks $\{ \by_t \}$ for each frame $t$, we produce a sequence of images $\{\bF_t\}$ representing the facial animation.
\rev{The input portrait might either depict a cartoon face, or a natural human face image}. We use different implementations for each of these two types of portraits. In the next paragraphs, we explain the variant used for each type.

\begin{figure}[t]
\begin{center}
\includegraphics[width=\columnwidth]{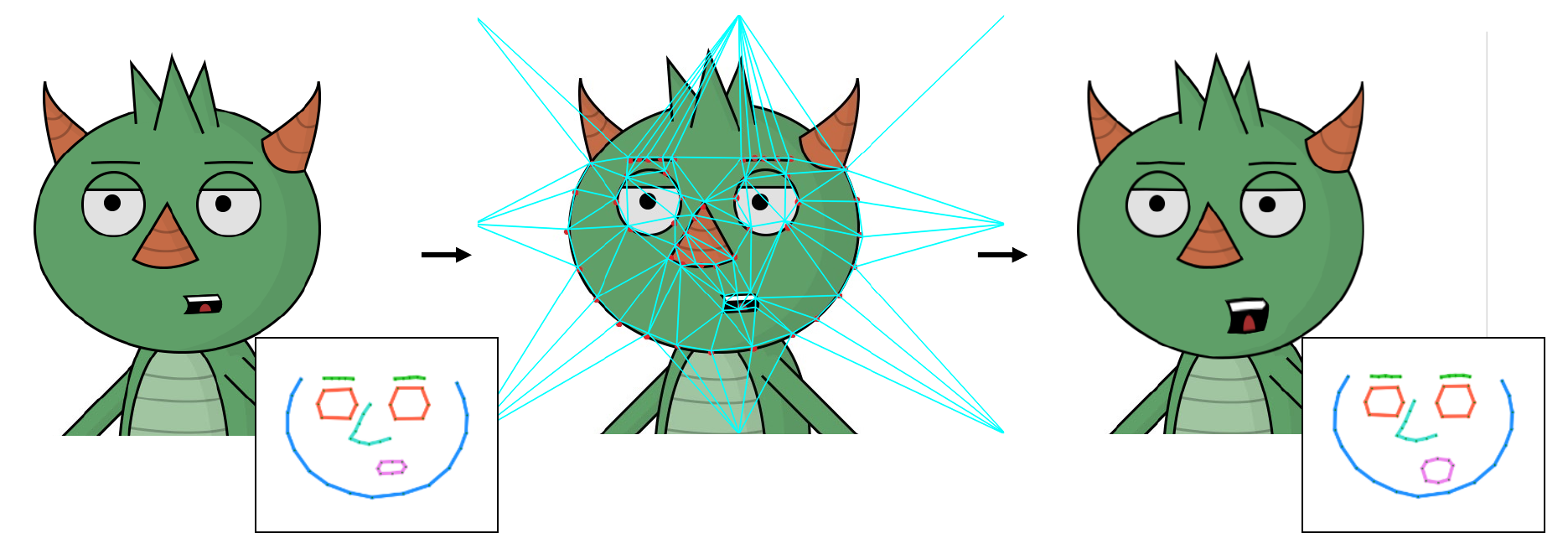}
\vspace{-6mm}
   \caption{Cartoon image face warping through facial landmarks and Delaunay Triangulation. Left: Given cartoon image and its facial landmarks. Middle: Delaunay triangulation. Right: Warped image guided by the displaced landmarks.}
\label{fig:delaunay}
\end{center}
\end{figure}

\paragraph{Cartoon Images (non-photorealistic).}
\label{sec:cartoon_image}
These images usually have sharp feature edges, e.g., from vector arts or flat shaded drawings.  
To preserve these sharp features, we propose a morphing-based method to animate them, avoiding pixel-level artifacts.
From the input image, we extract the facial landmarks using \cite{yaniv2019face}.
We then run Delaunay triangulation on these landmarks to generate semantic triangles.
By mapping the initial pixels as texture maps to the triangles, the subsequent animation process becomes straightforward. 
As long as the landmark topology remains the same, the textures on each triangle naturally transfer across frames. An illustration is shown in Figure~\ref{fig:delaunay}.
An analogy with our approach is the vertex and fragment shader pipeline in rendering.
The textures are bind to each fragment at the very beginning and from then on, only the vertex shader is changing the location of these vertices (landmark positions).
In practice, we implement a GLSL-based C++ code that uses vertex/fragment shaders and runs in real-time.

\rev{\paragraph{Natural Images.} 
The goal here is to synthesize a sequence of frames given the input photo and the predicted animated landmarks from our model. 
Our image synthesis approach is inspired by the landmark-based facial animation from \citet{zakharov2019few}, which translates landmarks to natural images based on a trained network. Instead of using a separate embedder and  adaptive instance normalization layers to encode the target face appearance, our module is designed based on the UNet architecture to process displaced landmarks and portraits. Specifically,} we first create an image representation $\bY_t$ of the predicted landmarks $\by_t$ by connecting consecutive facial landmarks and rendering them as line segments of predefined color (Figure~\ref{fig:overview}). 
The image $\bY_t$ is concatenated channel-wise with the input portrait image $\bQ$ to form a 6-channel image of resolution $256 \times 256$. The image is passed to an encoder-decoder network that performs image translation to produce the image $\bF_t$ per frame. Its architecture
follows the generators proposed in \citet{esser2018variational} and \citet{han2019finet}.
Specifically, the encoder employs 6 convolutional layers, where each layer contains one 2-strided convolution followed by two residual blocks, and produces a bottleneck, which is then decoded through symmetric upsampling decoders. Skip connections are utilized between symmetric layers of the encoder and decoder, as in U-net architectures \cite{ronneberger2015u}.
The generation proceeds for each frame.   
Since the landmarks change smoothly over time,  the output images formed as an interpolation of these landmarks exhibit  temporal coherence.
Examples of generated image sequences are shown in Figure~\ref{fig:natural_examples}.

\section{Training}
\label{sec:train}
We now describe our training procedure to learn the parameters of each module in our architecture.

\paragraph{Voice Conversion Training.}
We follow the training setup described in \cite{qian2019autovc} with the speaker embedding initialized by the pretrained model provided by \citet{wan2018resemblyzer}.  
A training source speech from each speaker is processed through the content encoder. Then another utterance of the same source speaker is used to extract the speaker embedding, which  is passed to the decoder
along with the audio content embedding
to reconstruct the original source speech.
The content encoder, decoder, and MLP are trained to minimize the self-reconstruction error of the source speech
spectrograms \cite{qian2019autovc}. Training is performed on the \emph{VCTK} corpus \cite{veaux2016superseded}, which is a speech dataset including utterances by 109 native English speakers with various accents.

\begin{figure}[t!]
\begin{center}
\includegraphics[width=1\linewidth]{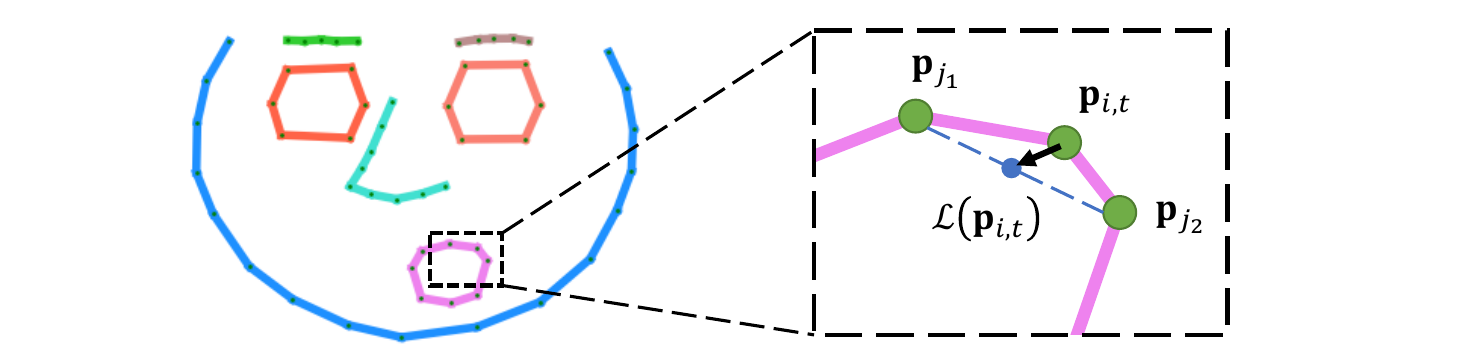}
\end{center}
\vspace{-3mm}
   \caption{Graph Laplacian coordinates illustration. Left: 8 facial parts that contain subsets of landmarks. Right: Zoom-in graph Laplacian vector and related neighboring landmark points.}
\label{fig:laplacian_smooth}
\end{figure}

\subsection{Speech Content Animation Training}

\paragraph{Dataset.} To train a content-based animation model, we use an audio-visual dataset that provides high-quality facial landmarks and corresponding audio.
To this end, we use the \textit{Obama Weekly Address} dataset \cite{suwajanakorn2017synthesizing} containing 6 hours of video featuring various Obama's speeches. 
Due to its high resolution and relatively consistent front facing camera angle, we can obtain accurate facial landmark detection results using~\cite{bulat2017far}. We also register the facial landmarks to a front-facing standard facial  template~\cite{beeler2014rigid} using a best-estimated affine transformation~\cite{segal2009generalized}. 
\rev{This also results in factoring out the speaker-dependent head pose motion.
We emphasize that one registered speaker is enough to train this module, since our goal here is to learn a generic lip motion based on audio. The lip sync will be specialized to particular speaker IDs by including the speaker-aware branch and training on multiple human subjects, as we will explain in \S\ref{sec:speaker-module-training}.}

\paragraph{Loss function.} To learn the parameters $\{\bw_{lstm,c}, \bw_{mlp,c}\}$ 
used in the LSTM and MLP,  we minimize a loss function that evaluates 
(a) the distance between the registered reference landmark positions $\hat{\bp}_t$ and predicted ones ${\bp}_t$, 
and (b) the distance between their respective graph Laplacian coordinates, which promotes correct placement of landmarks  with respect to each other and preserves facial shape  details~\cite{SorkineMeshProcessing06}.
Specifically, our loss is:
\begin{equation}
    L_{c} = \sum_{t=1}^T    \sum_{i=1}^N 
    \big\|  \bp_{i,t} - \hat{\bp}_{i,t}   \big\|_2^2
+  \lambda_c \sum_{t=1}^T     \sum_{i=1}^N 
   \big\| \mL( \bp_{i,t} ) - \mL(  \hat{\bp}_{i,t} ) \big\|_2^2,
\end{equation}
where $i$ is the index for each individual landmark,
and  $\lambda_c$ weighs the second term ($\lambda_c\!=\!1$ in our implementation, 
set through hold-out validation). 
We use the following graph Laplacian $\mL( \bp_t )$:
\begin{equation}
\mL( \bp_{i,t} )= 
   \bp_{i,t} - \frac{1}{|\mN(\bp_{i})|}  \sum_{\bp_j \in \mN(\bp_{i})} \bp_{j,t},
\label{eq:laplacian}
\end{equation} 
where $\mN(\bp_i)$ includes the  landmark neighbors connected to $\bp_i$ within a distinct facial part (Figure~\ref{fig:laplacian_smooth}). We use 8 facial parts that contain subsets of landmarks predefined for the facial template.

\subsection{Speaker-Aware Animation Training} 
\label{sec:speaker-module-training}

\paragraph{Dataset.} 
To learn the speaker-aware dynamics of head motion and facial expressions, we need an audio-visual dataset featuring a diverse set of speakers. 
We found the \emph{VoxCeleb2} dataset  is well-suited for our purpose since it contains video segments from a variety of speakers~\cite{Chung18b}. 
VoxCeleb2 was originally designed for speaker verification. 
Since our goal is different, i.e., capturing speaker dynamics for talking head synthesis, we chose a subset of 67 speakers with a total of 1,232 videos clips from VoxCeleb2. 
On average, we have around 5-10 minutes of videos for each speaker.
The criterion for selection was accurate landmark detection in the videos based on manual verification.
Speakers were selected based on Poisson disk sampling on the speaker representation space. We split this  dataset as 60\% / 20\% / 20\% for training, hold-out validation and testing respectively.
In contrast to the content animation step, we do not register the landmarks to a front-facing template since here we are interested in learning the overall head motion. 

\paragraph{Adversarial network.} Apart from capturing landmark position, we also aim to match the speaker's head motion and facial expression dynamics during training. To this end, we incorporate a GAN\ approach. Specifically, we create a discriminator network $Attn_d$ which follows a similar structure with the self-attention generator network in \S\ref{sec:speaker-module-training}. More details about its architecture are provided in the appendix.  The goal of the discriminator is to find out if the temporal dynamics of the speaker's facial landmarks look ``realistic'' or fake.\ It takes as input the sequence of facial landmarks within the same window used in the generator, along with audio content and speaker's embedding. It returns an output characterizing the ``realism'' $r_t$ per frame $t$:
\begin{equation}
r_t = {Attn}_d(\by_{t \rightarrow t+\tau'}, \tilde \bc_{t \rightarrow t+\tau'} , \bs; \bw_{\text{attn},d}),
\end{equation}
where $\bw_{\text{attn},d}$ are the parameters of the discriminator. We use the LSGAN loss function \cite{mao2017least} to train the discriminator parameters treating the training landmarks as ``real'' and the generated ones as ``fake'' for each frame:
\begin{equation}
    L_{gan} = \sum_{t=1}^T (\hat{r_t} - 1)^2 + r_t^2,
\end{equation}
where $\hat{r_t}$ denotes the discriminator output when the training landmarks $\hat{\by_t}$ are used as its input.

\paragraph{Loss function.} To train the parameters $\bw_{\text{attn},s}$ of the self-attention generator network, we attempt to maximize the ``realism'' of the output landmarks, and also consider the distance to the training ones in terms of absolute position and Laplacian coordinates: 
\begin{align}
    L_{s} &= \sum_{t=1}^T    \sum_{i=1}^N 
    \big\|  \by_{i,t} - \hat{\by}_{i,t}   \big\|_2^2
+  \lambda_s \sum_{t=1}^T     \sum_{i=1}^N 
   \big\| \mL( \by_{i,t} ) - \mL(  \hat{\by}_{i,t} ) \big\|_2^2   \nonumber  \\
& + \mu_s \sum_{t=1}^T (r_t - 1)^2,
\end{align}
where $\lambda_s=1$ and $\mu_s = 0.001$ are set through hold-out validation. We alternate training between the generator (minimizing $ L_{s}$) and discriminator (minimizing  $L_{gan}$) to improve each other as  done in GAN\ approaches \cite{mao2017least}.

\subsection{Image-to-Image Translation Training}

\rev{Finally, we train our image-to-image translation module to handle natural human face animation outputs.}
The encoder/decoder pair used for image translation is first trained on paired video frames from VoxCeleb2.
Then, we fine-tune the network on a subset which contains high-resolution video crops provided by \citet{Siarohin_2019_NeurIPS}.
In particular, based on a video of a talking person in the dataset, we randomly sample a frame pair: a source training frame $\hat{\bQ}_{src}$ and a target frame $\hat{\bQ}_{trg}$ of this person. 
The facial landmarks of the target face are extracted and rasterized into an RGB image $\hat\bY_{trg}$ based on the approach described in \S\ref{sec:animation}. 
The encoder/decoder network takes as input the concatenation of $\hat\bQ_{src}$ and $\hat\bY_{trg}$ and outputs a reconstructed   face $\bQ_{trg}$. 
The loss function aims to minimize the $L^1$ per-pixel distance and perceptual feature distance between the reconstructed  face $\bQ_{trg}$ and training target face $\hat\bQ_{trg}$ as in \cite{johnson2016perceptual}:
\begin{equation}
    L_{a} = \sum_{\{\src,\trg\}} || \bQ_{\trg} - \hat\bQ_{\trg}||_1 + 
    \lambda_a \!\!\!\!\!\! \sum_{\{\src,\trg\}}|| \phi(\bQ_{\trg}) - \phi(\hat\bQ_{\trg})||_1,
  \nonumber
\end{equation}
where $\lambda_a=1$, and $\phi$ concatenates feature map activations from the pretrained VGG19 network \cite{simonyan2014very}. 

\begin{figure*}[t!]
\begin{center}
\includegraphics[width=1\linewidth]{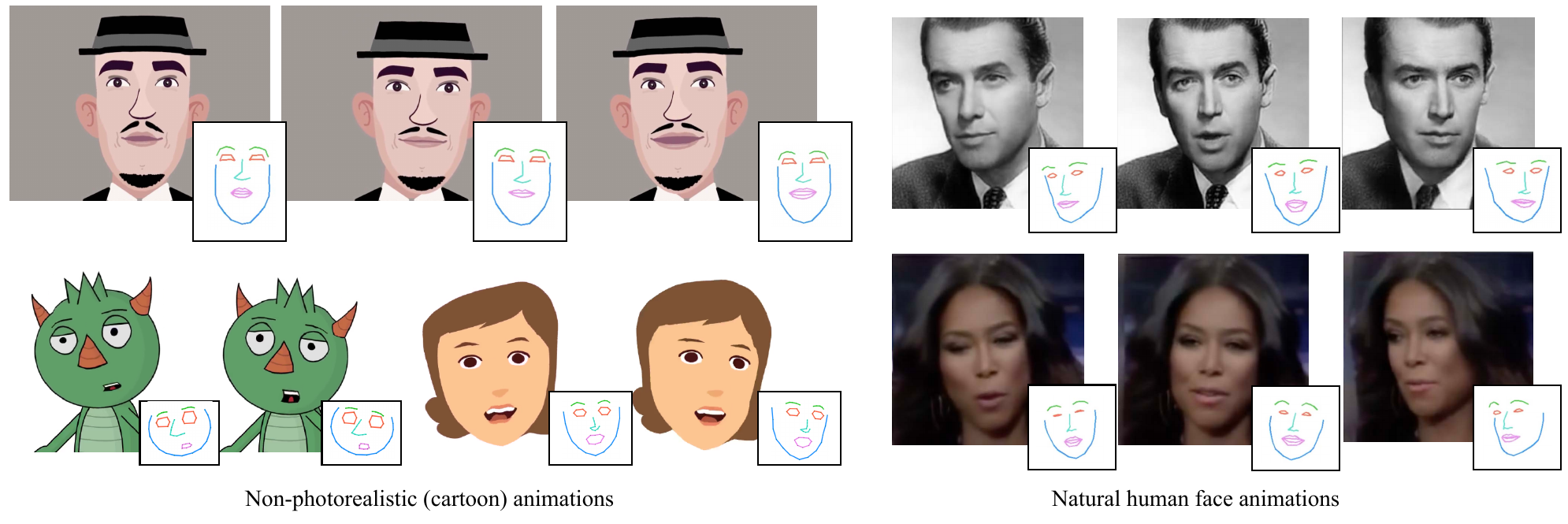}
\end{center}
\vspace{-4mm}
   \caption{Generated talking-head animation gallery for non-photorealistic cartoon faces (left) and \rev{natural human faces} (right). The corresponding intermediate facial landmark predictions are also shown on the right-bottom corner of each animation frame. Our method synthesizes not only facial expressions, but also different head poses. Cartoon \emph{Man with hat} and \emph{Girl with brown hair} \textcopyright \emph{Yang Zhou}. Natural face (at right bottom corner) from \emph{VoxCeleb2} dataset~\cite{Chung18b} \textcopyright \emph{Visual Geometry Group} (CC BY).}
\label{fig:head_pose}
\vspace{-1mm}
\end{figure*}

\subsection{Implementation Details}
All landmarks in our dataset are converted to $62.5$ frames per second and audio waveforms are sampled under $16K$ Hz frequency. Both of these rates followed \citet{qian2019autovc}, i.e. 62.5 Hz for the mel-spetrogram and 16 kHz for the speech waveform. 
We experimented with other common frame rates, and we found the above worked well for the entire pipeline. 
We note that the facial landmarks are extracted from the input video at its original frame rate and the interpolation is performed on landmarks rather than the original pixels.
We trained both the speech content animation and speaker-aware animation modules with the Adam optimizer using PyTorch.
The learning rate was set to $10^{-4}$, and weight decay to $10^{-6}$. 
The speech content animation module contains $1.9M$  parameters and took 12 hours to train on a Nvidia 1080Ti GPU. 
The speaker-aware  animation module has $3.8M$ parameters and took $30$  hours to train on the same GPU. The single-face animation module for generating natural human faces was also trained with the Adam optimizer, a learning rate of $10^{-4}$, and a batch size of $16$. 
The network has $30.7M$ parameters and was trained for $20$ hours on $8$ Nvidia 1080Ti GPUs. \rev{At test time, our network produces natural human face videos at around $22$ FPS, and animations for non-photorealistic videos at around $28$ FPS.}

\rev{Our source code and models are available at the following project page: \textcolor{blue}{https://people.umass.edu/~yangzhou/MakeItTalk/}
}

\section{Results}
\label{sec:result_natural}

\begin{figure*}[t!]
\begin{center}
\includegraphics[width=\linewidth]{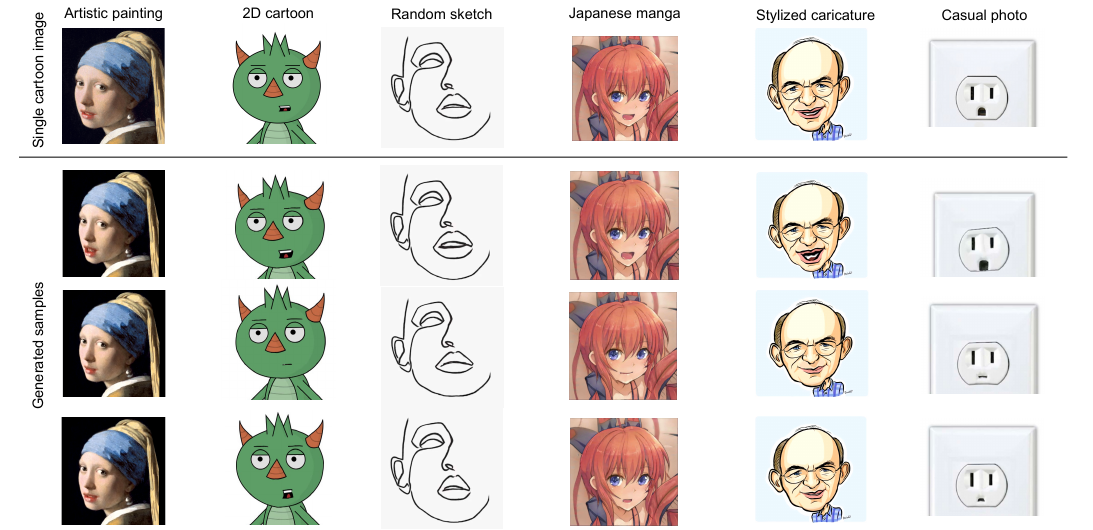}
\end{center}
\vspace{-4mm}
   \caption{Our model works for a variety types of non-photorealistic (cartoon) portrait images, including artistic paintings, 2D cartoon characters, random sketches, Japanese mangas, stylized caricatures and casual photos. Top row: input cartoon images. Next rows: generated talking face examples by face warping. Please also see our supplementary video. Artistic painting \emph{Girl with a pearl earring} \textcopyright \emph{Johannes Vermeer} (public domain). \emph{Random sketch} \textcopyright \emph{Yang Zhou}. \emph{Japanese manga} \textcopyright \emph{Gwern Branwen} (CC-0). Stylized caricature \textcopyright \emph{Daichi Ito} at Adobe Research.}
\label{fig:cartoon_examples}
\end{figure*}

\begin{figure*}[tb]
\begin{center}
\includegraphics[width=\linewidth]{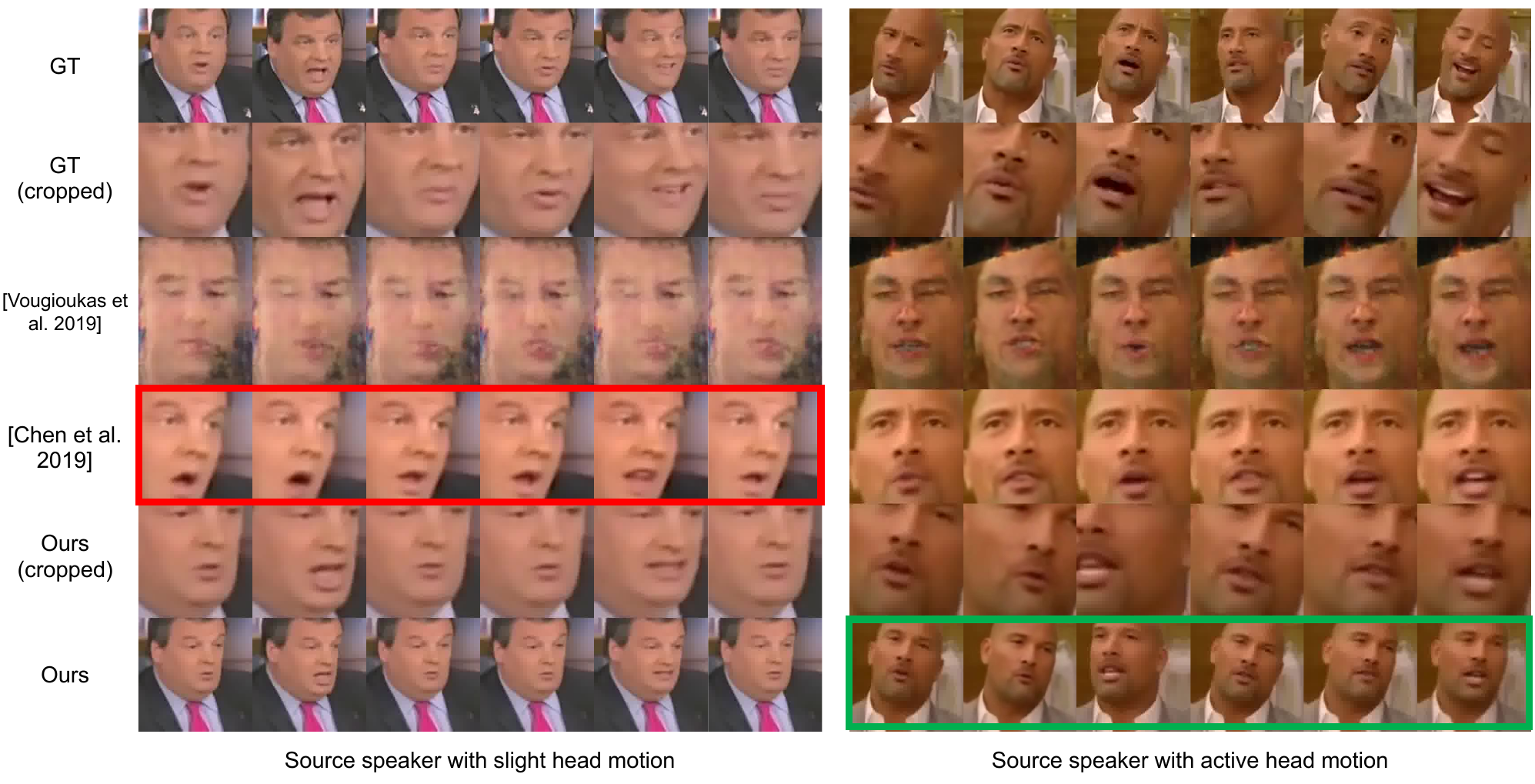}
\end{center}
\vspace{-3mm}
   \caption{Comparison with state-of-the-art methods for video generation of human talking-heads. 
   The compared methods crop the face and predict primarily the lip region while ours generates both facial expression and head motion. 
   GT and our results are full faces and are cropped for a better visualization of the lip region. 
   Left example:  \citet{chen2019hierarchical} has worse lip synchronization for side-faces (see the red box). 
   Right example: our method predicts speaker-aware head pose dynamics (see the green box). Note that the predicted head pose is different than the one in the ground-truth video, but it exhibits similar \emph{dynamics} that are characteristic for the speaker. Natural faces from \emph{VoxCeleb2} dataset~\cite{Chung18b} \textcopyright \emph{Visual Geometry Group} (CC BY). }
\label{fig:natural_examples}
\end{figure*}

With all the pieces in place, we now present results of talking-head videos from a single input image and a given audio file. 
Figure~\ref{fig:head_pose} shows a gallery of our generated animations for cartoon and \rev{natural human face images}.  
We note that the resulting animations include both full facial expressions and dynamic head poses. 
We refer readers to our supplementary video.

\rev{In the following sections, we discuss more results for generating non-photorealistic animations, and natural human facial video}, along with qualitative comparisons. Then we present detailed numerical evaluation, an ablation study, and applications of our method.

\subsection{Animating Non-Photorealistic Images}
Figure~\ref{fig:cartoon_examples} shows a gallery of our generated non-photorealistic animations. Each animation is generated from input audio and a single portrait image. The portrait images can be   artistic paintings, random sketches, 2D cartoon characters,  Japanese mangas, stylized caricatures and casual photos.
Despite being only trained on human facial landmarks, our method can successfully generalize to a large variety of stylized cartoon faces. 
This is because our method uses landmarks as intermediate representations, and also learns relative landmark displacements instead of absolute positions.

\subsection{Animating Human Facial Images} 
\rev{Figure~\ref{fig:natural_examples} shows
synthesized talking head videos featuring talking people} as well as comparisons with state-of-the-art video generation methods
~\cite{chen2019hierarchical, vougioukas2019realistic}. 
The ground-truth (GT) and our results are cropped to highlight the differences in the lip region (see row 2 and 5).
We notice that the videos generated by  \citet{vougioukas2019realistic, chen2019hierarchical} predict primarily the lip region on cropped faces and therefore miss the head poses.
\citet{vougioukas2019realistic} does not generalize well to faces unseen during training.
\citet{chen2019hierarchical} lacks synchronization with the input audio, especially for side-facing portraits (see the red box). 
Compared to these methods, our method predicts facial expressions more accurately and also captures head motion to some degree (see the green box). On the other hand, to be fair, we also note that our method is not artifact-free: the head motion often distorts the background, which is due to the fact that our network attempts to produce the whole image, without separating foreground from background.

Even only trained on natural face images, our image translation module can also generate plausible facial animations not only limited to real faces, but also to 2D paintings, picture of statue heads, or rendered images of 3D models. Please check our supplementary video for results.

Our supplementary video also includes a comparison with the concurrent work by \cite{thies2019neural}. Given the same audio and target speaker image at test time, we found that our lip synchronization appears to be more accurate than their method. We also emphasize that our method learns to generate head poses, while in their case the head pose is not explicitly handled or is added back heuristically in a post-processing step (not detailed in their paper).
Their synthesized video frames appear sharper than ours perhaps due to their neural renderer of their 3D face model. \rev{However, their method requires additional training on target-specific reference videos of length around 30 hours, while ours animates a single target photo immediately without any retraining. 
Thus, our network has the distinctive advantage of driving a diverse set of single images for which long training videos are not available. These include static cartoon characters, casual photos, paintings, and sketches.}

\subsection {Evaluation Protocol}
We evaluated MakeItTalk and compared with related methods quantitatively and with user studies. 
We created a test split from the VoxCeleb2 subset, containing $268$ video segments from $67$ speakers.
The speaker identities were observed during training, however, their test speech and video are different from the training ones.
Each video clip lasts $5$ to $30$ seconds. 
Landmarks were extracted using~\cite{bulat2017far} from test clips and their quality was manually verified. 
We call these as ``reference landmarks'' and use them in the evaluation metrics explained below.


\paragraph{Evaluation Metrics.} To evaluate how well the synthesized landmarks represent accurate lip movements, we use the following metrics:

\begin{itemize}[leftmargin=5mm]
    \item \textbf{Landmark distance for jaw-lips (D-LL)} represents the average Euclidean distance between predicted facial landmark locations  of the jaw and lips and reference ones. The landmark positions are normalized according to the maximum width of the reference lips for each test video clip.
    \item \textbf{Landmark velocity difference for jaw-lips (D-VL)} represents the average Euclidean distance between reference landmark velocities of the jaw and lips and  predicted ones. Velocity is computed as the difference of landmark locations between consecutive frames. The metric captures differences in  first-order jaw-lips dynamics.
    \item \textbf{Difference in open mouth area  (D-A:)}: the average difference between the area of the predicted mouth shape and reference one. It is  expressed as percentage of the maximum area of the  reference mouth for each test video clip.

\end{itemize}

To evaluate how well the landmarks produced by our method and others reproduce overall head motion, facial expressions, and their dynamics, we use the following metrics:

\begin{itemize}[leftmargin=5mm]
  \item \textbf{Landmark distance (D-L)}: the average Euclidean distance between all predicted  facial landmark locations and reference ones (normalized by the width of the face).
    \item \textbf{Landmark velocity difference (D-V)}: the average Euclidean distance between reference landmark velocities and  predicted ones (again normalized by the width of the face). Velocity is computed as the difference of landmark locations between consecutive frames. This metric serves as an indicator of landmark motion dynamics.
    \item \textbf{Head rotation and position difference (D-Rot/Pos)}: the average difference between the reference and predicted head rotation angles (measured in degrees) and head position (again normalized by the width of the face). The measure indicates head pose differences, like nods and tilts.

\end{itemize}

\begin{table}
\caption{Quantitative comparison of facial  landmark predictions of MakeItTalk versus  state-of-the-art methods.}
\vspace{-3mm}
\begin{center}
\begin{tabular}{c|c|c|c}
    Methods & D-LL $\downarrow$ & D-VL $\downarrow$ & D-A $\downarrow$ \\ \hline\hline
    \cite{zhou2018visemenet} & 6.2\% & 0.63\% & 15.2\% \\ \hline
    \cite{eskimez2018generating} & 4.0\% & 0.42\% & 7.5\% \\ \hline
    \cite{chen2019hierarchical} & 5.0\% & 0.41\%  & 5.0\% \\ \hline\hline
    \rev{Ours (no separation)} & 2.9\% & 0.64\% & 17.1\% \\ \hline
    Ours (no speaker branch) & 2.2\% & 0.29\% & 5.9\% \\ \hline
    Ours (no content branch) & 3.1\% & 0.38\% & 10.2\% \\ \hline
    Ours (full) & \textbf{2.0\%} & \textbf{0.27\%} & \textbf{4.2\%}
\end{tabular}{}
\end{center}
\label{table:compare_landmark_with_others}
\end{table}

\subsection{Content Animation Evaluation}
\label{sec:eval_content}

We begin our evaluation by comparing MakeItTalk with state-of-the-art methods for synthesis of facial expressions driven by landmarks. 
Specifically, we compare with  \citet{eskimez2018generating}, \citet{zhou2018visemenet}, and \citet{chen2019hierarchical}. All these methods  attempt to synthesize facial landmarks, but cannot produce  head motion. 
Head movements are either generated procedurally or copied from a source video. 
Thus, to perform a fair evaluation, we factor out head motion from our method, and focus only on comparing predicted landmarks under an identical ``neutral'' head pose for all methods.
For the purpose of this evaluation, we focus on the lip synchronization metrics (\textbf{D-LL}, \textbf{D-VL}, \textbf{D-A}), and ignore head pose metrics. 
Quantitatively, Table~\ref{table:compare_landmark_with_others} reports these metrics for the above-mentioned methods and ours. 
\rev{We include our full method, including three reduced variants: (a) \emph{``Ours (no separation)''}, where we eliminate the voice conversion module and feed the raw audio features as input directly to the speaker-aware animation branch trained and tested alone; in this manner, there is no separation (disentanglement) between audio content and speaker identity,
(b) \emph{``Ours (no speaker branch)''}, where we keep the voice conversion module for disentanglement, but we train and test the speech content branch alone without the speaker-aware branch, (c) \emph{``Ours (no content branch)''}, where we again perform disentanglement, but we train and test the speaker-aware branch alone without the speech content branch. We discuss these three variants in more detail in our ablation study (Section \ref{sec:ablation_study}).}
The result shows that our method achieves the lowest errors for all measures. In particular, our method has $2$x times less \textbf{D-LL} error in lip landmark positions compared to \cite{eskimez2018generating}, and $2.5$x times less \textbf{D-LL} error compared to \cite{chen2019hierarchical}.

\begin{figure}
\begin{center}
\includegraphics[width=\linewidth]{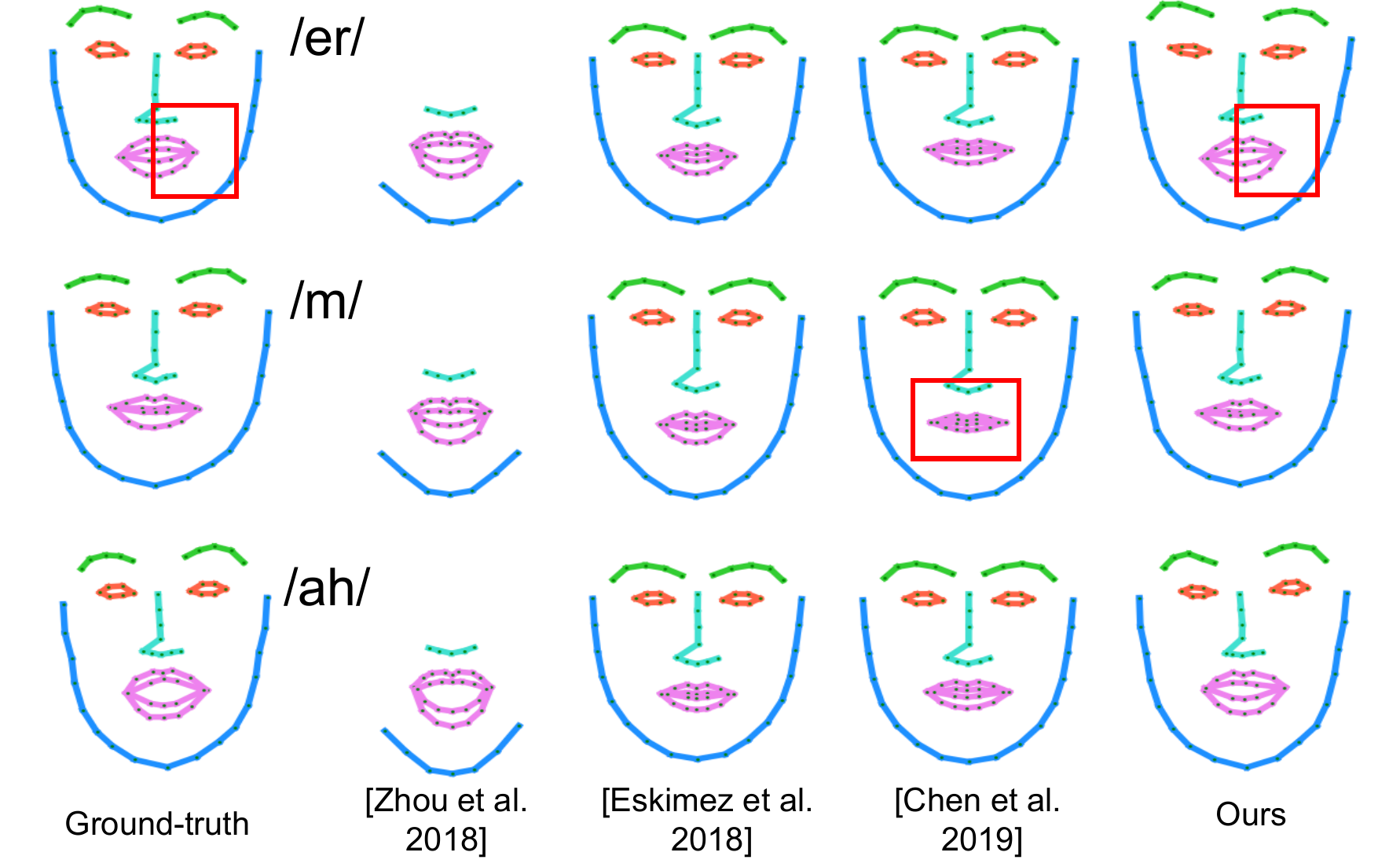}
\end{center}
\vspace{-3mm}
   \caption{Facial expression landmark comparison. Each row shows an example frame prediction for different methods. The GT landmark and uttered phonemes are shown on left. 
   }
\label{fig:compare_landmark_with_others}
\vspace{-2mm}
\end{figure}

Figure~\ref{fig:compare_landmark_with_others} shows characteristic examples of  facial landmark outputs for the above methods and ours from our test set. 
Each row shows one output frame. 
\citet{zhou2018visemenet} is only able to  predict the lower part of the face and cannot reproduce closed mouths accurately (see second row).  
\citet{eskimez2018generating} and \citet{chen2019hierarchical}, on the other hand, tend to favor conservative mouth opening. 
In particular, \citet{chen2019hierarchical}  predicts bottom and upper lips that sometimes  overlap with each other (see second row, red box). 
In contrast, our method captures facial expressions that match better the reference ones. 
Ours can also predict subtle facial expressions, such as the lip-corner lifting (see first row, red box).

\subsection{Speaker-Aware Animation Evaluation}
\label{sec:eval_style}

\paragraph{Head Pose Prediction and Speaker Awareness}

Existing speech-driven facial animation methods do not synthesize head motion. 
Instead, a common strategy is to copy head poses from another existing video. 
Based on this observation, we evaluate our method against two baselines: \emph{``retrieve-same ID''} and \emph{``retrieve-random ID''}. These baselines retrieve the head pose and position sequence from another video clip randomly picked from our training set. Then the facial landmarks are translated and rotated to reproduce the copied head poses and positions. The first baseline \emph{``retrieve-same ID''} uses a training video with the same speaker as in the test video. This strategy makes this baseline stronger since it re-uses dynamics from the same speaker. The second baseline \emph{``retrieve-random ID''} uses a video from a different random speaker. This baseline is useful to examine whether our method and alternatives produce head pose and facial expressions better than random or not.

Table ~\ref{table:compare_head_pos} reports the \textbf{D-L},  \textbf{D-V}, and \textbf{D-Rot/Pos} metrics.
Our full method achieves much smaller errors compared to both baselines, indicating our speaker-aware prediction is more faithful compared to  merely copying head motion from another video. In particular, we observe that our method produces $2.7x$ less error in head pose (D-Rot), and  $1.7x$ less error in head position (D-Pos) compared to 
using a random speaker identity (see \emph{``retrieve-random ID''}). This result also confirms that the head motion dynamics of random speakers largely differ from ground-truth ones. 
Compared to the stronger baseline of re-using video from the same speaker (see \emph{``retrieve-same ID''}), we observe that our method still produces  $1.3x$ less error in head pose (D-Rot), and  $1.5x$ less error in head position (D-Pos). This result confirms that re-using head motion from a video clip even from the right speaker still results in significant discrepancies, since the copied head pose and position does not necessarily synchronize well with the audio. Our full method instead captures the head motion dynamics and facial expressions more consistently w.r.t. the input audio and speaker identity.

\begin{figure}[t!]
\begin{center}
\includegraphics[width=0.95\linewidth]{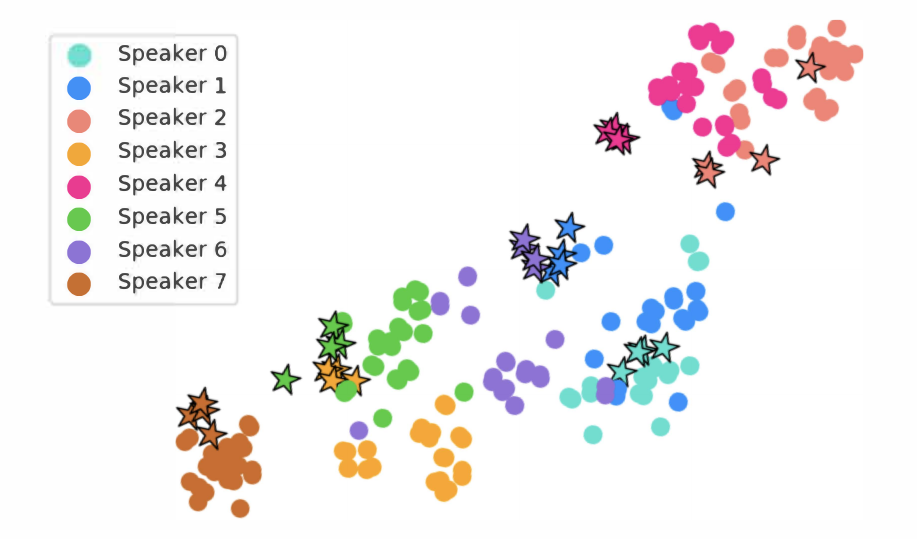}
\end{center}
\vspace{-5mm}
   \caption{t-SNE visualization for AUs, head pose and position variance based on 8 reference speakers videos (solid dots) and our predictions (stars). Different speakers are marked with different colors as shown in the legend.}
\vspace{-2mm}
\label{fig:tsne_aus}
\end{figure}

Figure~\ref{fig:head_pose} shows a gallery of our generated cartoon images and natural human faces under different predicted head poses. 
The corresponding generated facial landmarks are also shown on the right-bottom corner of each image. The demonstrated examples  show that our method is able to synthesize head pose well, including nods and swings. 
Figure~\ref{fig:tsne_aus} shows another qualitative validation of our method's ability to capture personalized head motion dynamics.
The figure embeds $8$ representative speakers from our dataset based on their variance in Action Units (AUs), head pose
and position variance. The 
AUs are computed from the predicted landmarks based on the definitions from \cite{friesen1978facial}.
The embedding is performed through t-SNE~\cite{maaten2008visualizing}. 
These $8$ representatives were selected using furthest sampling i.e., their AUs, head pose and position differ most from the rest of the speakers in our dataset.
We use different colors for different speakers and use \emph{solid dots} for embeddings produced based on the reference videos in AUs, head pose and position variance, and  \emph{stars} for embeddings resulting from our method.   
The visualization demonstrates that our method produces head motion dynamics that tend to be located more closely to reference ones.

\begin{table}
\caption{Head pose prediction comparison with the baseline methods in \S\ref{sec:eval_style} and our variants based on our head pose metrics.}
\begin{center}
\vspace{-3mm}
\begin{tabular}{c|c|c|c}
    Methods & D-L $\downarrow$ & D-V $\downarrow$ & D-Rot/Pos $\downarrow$ \\ \hline\hline
    retrieve-same ID & 17.1\% & 1.2\% & 10.3/8.1\%  \\ \hline
    retrieve-random ID & 20.8\% & 1.1\% & 21.4/9.2\% \\ \hline
    \hline
    \rev{Ours (no separation)}  & 12.4\% & 1.1\% & 8.8/5.4\% \\ \hline
    Ours (random ID) & 33.0\% & 2.4\% & 28.7/12.3\% \\ \hline
    Ours (no speaker branch) & 13.8\% & 1.2\% & 12.6/6.9\% \\ \hline
    Ours (no content branch) & 12.5\% & 0.9\% & 8.6/5.7\% \\ \hline
    Ours (full) & \textbf{12.3\%} & \textbf{0.8\%} & \textbf{8.0/5.4\%} 
\end{tabular}{}
\end{center}
\label{table:compare_head_pos}
\end{table}

\subsection{Ablation study}
\label{sec:ablation_study}

\paragraph{Individual branch performance.}
\rev{We performed an ablation study by training and testing the three reduced variants of our network described in \S\ref{sec:eval_content}: \emph{``Ours (no separation)''} (no disentanglement between content and speaker identity), \emph{``Ours (no speaker branch)''}, and \emph{``Ours (no content branch)''}. The aim of the last two variants is to check whether a single network can jointly learn both lip synchronization and speaker-aware head motion.}
The results of these three variants and our full method are shown in in Table \ref{table:compare_landmark_with_others} and Table \ref{table:compare_head_pos}. We also refer readers to the supplementary video for more visual comparisons.

The variant \emph{``Ours (no speaker branch)''} only predicts facial landmarks from the audio content without considering the speaker identity. It performs well in terms of capturing the lip landmarks, since these are synchronized with the audio content. The variant  is slightly worse than our method based on the lip evaluation metrics (see Table \ref{table:compare_landmark_with_others}). However, it results in $1.6x$
larger errors in head pose and $1.3x$ larger error in head position
(see Table~\ref{table:compare_head_pos}) since head motion is a function of both speaker identity and content.

\begin{figure}[tb]
\begin{center}
\includegraphics[width=1\linewidth]{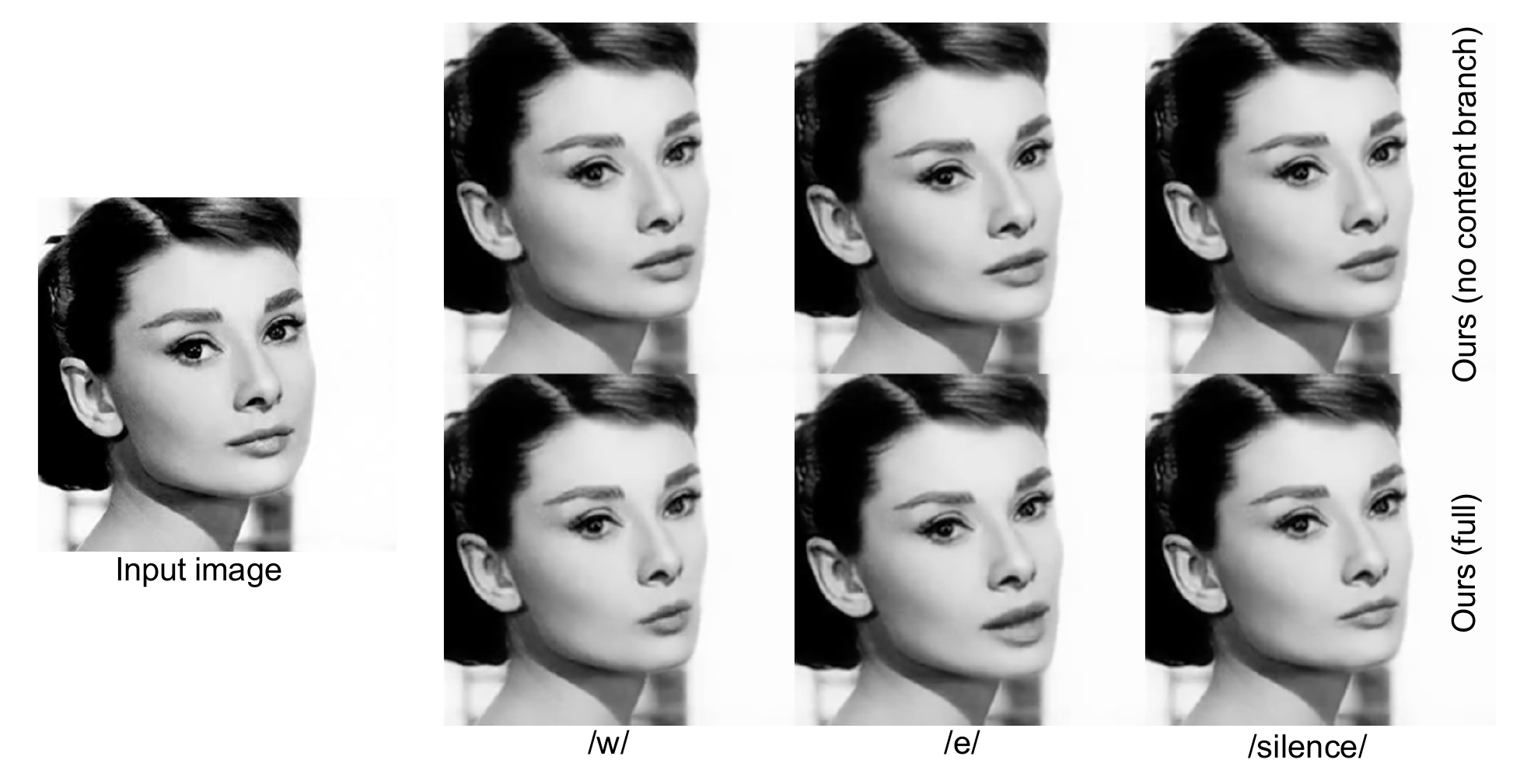}
\end{center}
\vspace{-3mm}
   \caption{Comparison to \emph{``Ours (no content branch)''} variant (right-top) which uses only the \emph{speaker-aware animation} branch. The full model (right-bottom) result has much better articulation in the lower-part of the face. It demonstrates that a single network architecture cannot jointly learn both lip synchronization and speaker-aware head motion. \emph{Audrey Hepburn} \textcopyright \emph{Me Pixels} (CC-0).}
\label{fig:ours_vs_spkonly}
\vspace{-3mm}
\end{figure}

The results of the variant \emph{``Ours (no content branch)''} has the opposite behaviour: it performs well in terms of capturing head pose and position
(it is slightly worse than our method, see Table~\ref{table:compare_head_pos}). However, it has $1.6x$ higher error in jaw-lip landmark difference and $2.4x$ higher error in open mouth area difference (see Table~\ref{table:compare_landmark_with_others}), which indicates that the lower part of face dynamics are not synchronized well with the audio content. Figure \ref{fig:ours_vs_spkonly} demonstrates  that using the speaker-aware animation branch alone i.e., the ``Ours (no content)'' variant results in noticeable artifacts in the jaw-lip landmark displacements. 
Using both branches in our full method offers the best performance according to all evaluation metrics.

\rev{The results of the variant \emph{``Ours (no separation)''} are similar to the variant \emph{``Ours (no content branch)''}: it 
achieves slightly worse head pose performance than our full method (Table~\ref{table:compare_head_pos}), and much worse results in terms of lip movement accuracy (Table~\ref{table:compare_landmark_with_others}). Specifically, it has $1.5x$, $2.4x$, and $4.1x$ higher error in jaw-lip landmark position, velocity, and open mouth area difference respectively. We hypothesize this is because the content and the speaker identity information are still entangled and therefore it is hard for the network to disambiguate a one-to-one mapping between audio and landmarks.} 

\paragraph{Random speaker ID injection} We tested one more variant of our method called \emph{``Ours (random ID)''}. For this variant, we use our full network, however, instead of using the correct speaker embedding, we inject another random speaker identity embedding. The result of this variant is shown in Table 3. Again we observe that the performance is significantly worse ($3.6$x more error for head pose). This indicates that our method successfully splits the content and speaker-aware motion dynamics, and captures the correct speaker head motion dynamics (i.e., it does not reproduce random ones).

\subsection{User Studies}
We also evaluated our method through perceptual user studies via Amazon Mechanical Turk service. 
We obtained $6480$ query responses from $324$ different MTurk participants in our two different user studies described below.

\begin{figure}[!t]
\begin{center}
\includegraphics[width=1\linewidth]{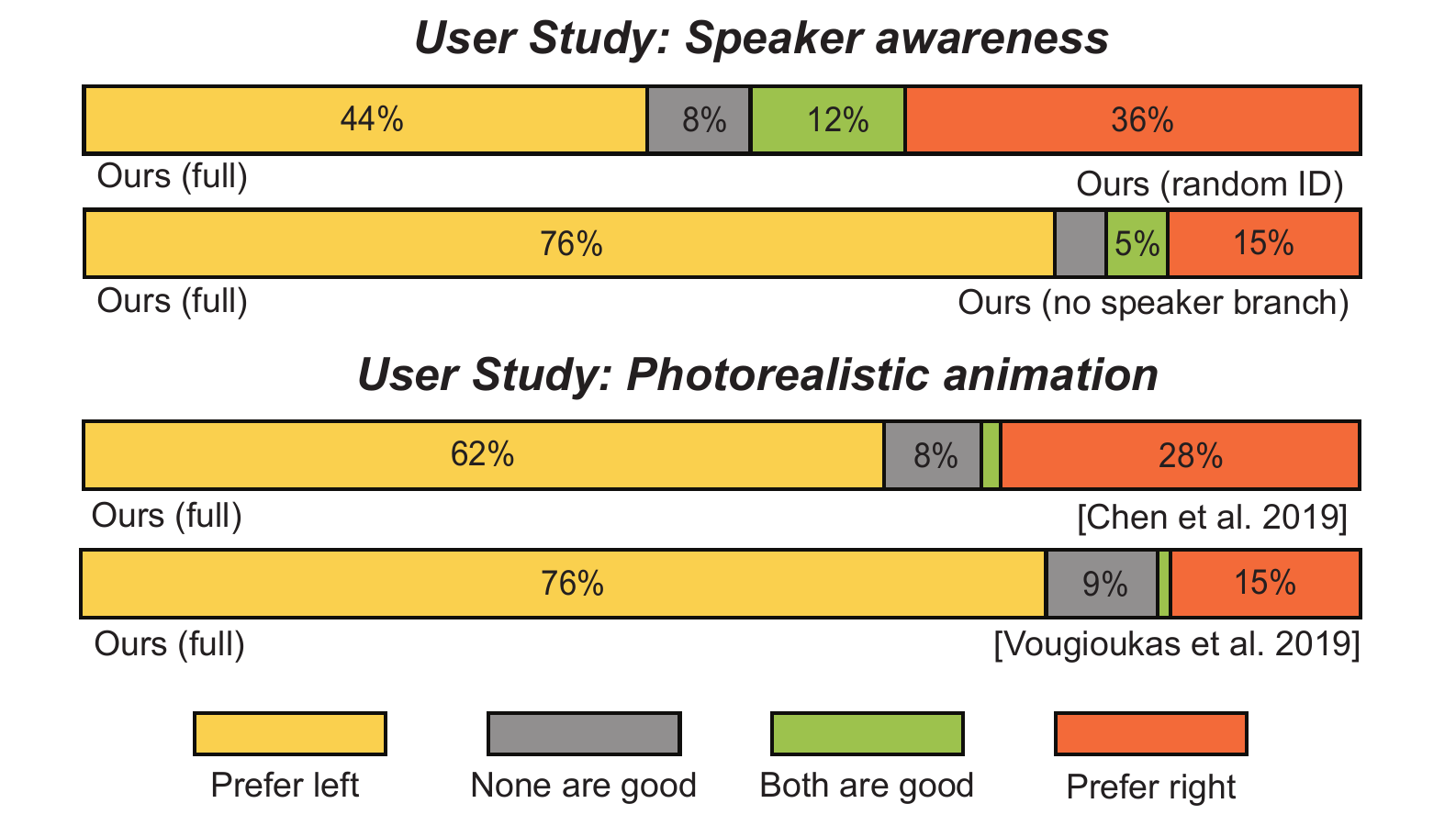}
\end{center}
\vspace{-4mm}
   \caption{User study results for speaker awareness (top) and natural human facial animation (bottom).}
\label{fig:user_study}
\vspace{-5mm}
\end{figure}

\paragraph{User study for speaker awareness}
\rev{Our first study evaluated the \emph{speaker awareness} of different variants of our method while synthesizing cartoon animations. Specifically,} we assembled a pool of $300$ queries displayed on different webpages. On top of the webpage, we showed a  reference video of a real person talking, and on the bottom we showed two cartoon animations: one cartoon animation generated using our full method and another cartoon animation based on one of the two variants discussed above: (\emph{``Ours (random ID)''} and \emph{``Ours (no speaker ID)''}.
The two cartoon videos were placed in randomly picked left/right positions for each webpage.
Each query included the following question:
``Suppose that you want to see the real person of the video on the top
  as the cartoon character shown on the bottom. 
  Which of the two cartoon animations best represents the person's
  talking style in terms of facial expressions and head motion?''
The MTurk participants were asked to pick one of the following choices: ``left animation'', ``right animation'', ``can't tell - both represent the person quite well'',  ``can't tell - none represent the person well''.  Each MTurk participant was asked to complete a questionnaire with $20$ queries randomly picked from our pool. Queries were shown at a random order. Each query was repeated twice (i.e., we had $10$ unique queries per questionnaire), with the two cartoon videos randomly flipped each time to detect unreliable participants giving inconsistent answers. 
 We filtered out unreliable MTurk participants who gave two different answers to more than $5$ out of the $10$ unique queries in the questionnaire, or took less than a minute to complete it.
Each participant was allowed to answer one questionnaire maximum to ensure participant diversity. 
We had $90$ different, reliable MTurk participants for this user study.
For each of our $300$ queries, we got votes from $3$ different MTurk participants. Since each MTurk participant voted for $10$ unique queries twice, we  gathered $1800$ responses ($300$ queries $\times$ 3 votes $\times$ 2 repetitions) from our $90$ MTurk participants. Figure~\ref{fig:user_study} (top) shows the study result. 
We see that the majority of MTurkers picked our full method more frequently, when compared with either of the two variants.

\rev{\paragraph{User study for natural human facial video.}
To validate our landmark-driven} \emph{human facial animation} method, we conducted one more user study. 
Each MTurk participant was shown a questionnaire with $20$ queries involving random pairwise comparisons out of a pool with $780$ queries we generated.
For each query, we showed a single frame showing the head of a real person on top, and two generated videos below (randomly placed at left/right positions): one video synthesized from our method, and another from either~\citet{vougioukas2019realistic} or  \citet{chen2019hierarchical}. The participants were asked which person's facial expression and head motion look more realistic and plausible. 
We also explicitly instruct them to ignore the particular camera position or zoom factor and focus on the face.
Participants were asked to pick one of four choices (``left'', ``right'', ``both'' or ``none'') as in the previous study. 
We also employed the same random and repeated query design and MTurker consistency and reliability checks to filter out unreliable answers. 
We had $234$ different MTurk participants for this user study. Like in the previous study, each query received votes from $3$ different, reliable MTurk participants. As a result, we gathered $780$ queries $\times$ $3$     votes $\times$ $2$ repetitions
= $4680$ responses from our  $234$ participants.
Figure~\ref{fig:user_study} (bottom) shows the study result. 
Our method was voted as the most ``realistic'' and ``plausible'' by a large majority, when compared to  \citet{chen2019hierarchical} or \citet{vougioukas2019realistic}.

\subsection{Applications}
\begin{figure}[t!]
\begin{center}
\includegraphics[width=\linewidth]{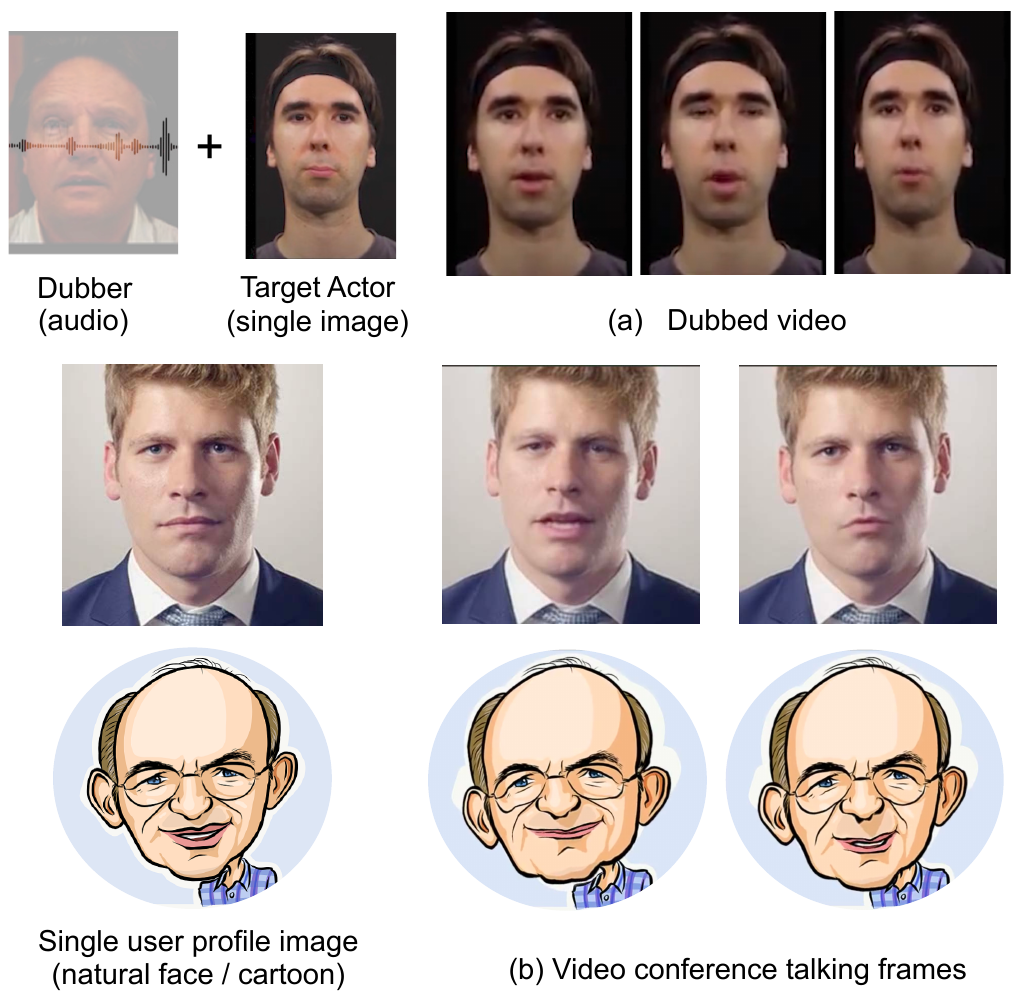}
\end{center}
\vspace{-5mm}
   \caption{Applications. Top row: video dubbing for target actor given only audio as input. Middle and bottom row: video conference for natural human and cartoon user profile images. Please also see our supplementary video. Video conference application \emph{natural face} \textcopyright \emph{PxHere} (CC-0).}
\vspace{-3mm}   
\label{fig:dubbing}
\end{figure}

\rev{Synthesizing plausible talking-heads has numerous applications} \cite{kim2019neural,zhou2019talking}.
One common application is dubbing using voice from a person different from the one in the original video, or even using voice in a different language. 
In Figure~\ref{fig:dubbing}(top), given a single frame of the target actor and a dubbing audio spoken by another person,
we can generate a video of the target actor talking according to that other person's speech.

Another application is bandwidth-limited video conference. 
In scenarios where the visual frames cannot be delivered with high fidelity and frame-rate, we can make use only of the audio signal to drive the talking-head video. 
Audio signal can be preserved under much lower bandwidth compared to its visual counterpart. Yet, it is still important to preserve visual facial expressions, especially lip motions, since they heavily contribute to understanding in communication~\cite{mcgurk1976hearing}.
Figure~\ref{fig:dubbing}(middle) shows that we can synthesize talking heads with facial expressions and lip motions with only the audio and an initial high-quality user profile image as input. Figure~\ref{fig:dubbing}(bottom) shows examples of both natural human and cartoon talking-head animation that can be used in teleconferencing for entertainment reasons, or due to privacy concerns related to video recording. We also refer readers to the supplementary video.

Our supplementary video also demonstrates a  text-to-video application, where we synthesize natural human face video from text input, after converting it to audio through a speech synthesizer \cite{notevibes}. Finally, our video demonstrates the possibility of interactively editing the pose of our synthesized talking heads by applying a rotation to the intermediate landmarks predicted by our network.

\section{Conclusion}
We have introduced a deep learning based approach to generate speaker-aware talking-head animations from an audio clip and a single image. 
Our method can handle new audio clips and new portrait images not seen during training.
Our key insight was to predict landmarks from disentangled audio content and speaker, such that they capture better lip synchronization, personalized facial expressions and head motion dynamics. This led to much more expressive animations with higher overall quality compared to the state-of-the-art. 

\paragraph{Limitations and Future Work}
There are still several avenues for future research. 
Although our method captures aspects of the speaker's style e.g., predicting head swings reflecting active speech, there are several other factors that can influence head motion dynamics. For example, the speaker's mood can also play a significant role in determining head motion and facial expressions.
Further incorporating sentiment analysis into the animation pipeline is a promising research direction.

Our speech content animation currently does not always capture well
the bilabial and fricative sounds, i.e. /b/m/p/f/v/. 
We believe this is caused by the voice conversion module that we used, which tends to miss those short phoneme representations when performing voice spectrum reconstruction.
While we observe promising results via directly adopting a state-of-the-art voice conversion architecture, a more domain-specific adaptation for audio-driven animation may address such discrepancies between voice conversion and animation.

\revrev{Improving the image translation from landmarks to videos can also be further investigated. 
The current image-to-image translation network takes only the 2D facial landmarks as input. 
Incorporating more phoneme- or viseme-related features as input may improve the quality of the generated video in terms of articulation. 
Moreover, background distortion and artifacts are noticeable in our current solution.
Our image translation module warps both the background and the foreground head to produce the animation, which gives an impression of a camera motion mixed with head motion. 
Adapting fore/background separation or portrait matting in the network architecture during training and testing may help generate better results~\cite{BMSengupta20}.}
Capturing long-range temporal and spatial dependencies between pixels could further reduce low-level artifacts. 
\revrev{Another limitation is that in the case of large head motion, more artifacts tend to appear: since we attempt to create animations from a single input image, large rotations/translations require sufficient extrapolation to unseen parts of the head (e.g., neck, shoulders, hair),  which are more challenging for our current image translation net to hallucinate especially for natural head images.}



\revrev{Our method heavily relies on the intermediate sparse landmark representation to guide the final video output. The representation has the advantage of being low-dimensional and handling a variety of faces beyond human-looking ones. On the other hand, landmarks serve mostly as coarse proxies for modeling heads, thus, for large motion, they sometimes cause face distortions especially for natural head images (see also our supplementary video, last clip). In the particular case of human face animation, an alternative representation could be denser landmarks or parameters of a morphable model that may  result in more accurate face reconstructions. A particular challenge here would be to train such models in the zero-shot learning regime, where the input portrait has not been observed during training; current methods seem to require additional fine-tuning on target faces \cite{thies2019neural}.
}


Finally, our current algorithm focuses on a fully automatic pipeline.
It remains an open challenge to incorporate user interaction within a human-in-the-loop approach. 
An important question is how an animator could edit landmarks in certain frames and propagate those edits to the rest of the video.
We look forward to future endeavors on high-quality expressive talking-head animations with intuitive controls.

\section{Ethical Considerations}

``Deepfake videos'' are becoming more prevalent in our everyday life. The general public might still think that talking head videos are hard or impossible to generate synthetically. As a result, algorithms for talking head generation can be misused to spread misinformation or for other malicious acts.  We hope our code will help people understand that generating such videos is entirely feasible. Our main intention is to spread awareness and demystify this technology. Our code includes a watermark to the generated videos making it clear that they are synthetic.

\begin{acks}
We would like to thank Timothy Langlois for the narration, and Kaizhi Qian for the help with the voice conversion module. We thank Jakub Fišer for implementing the real-time GPU version of the triangle morphing algorithm.
We thank Daichi Ito for sharing the caricature image and Dave Werner for Wilk, the gruff but ultimately lovable puppet.
We also thank the anonymous reviewers for their constructive comments and suggestions. 
This research is partially funded by NSF (EAGER-1942069) and a gift from Adobe.
Our  experiments  were  performed  in  the UMass GPU cluster obtained under the Collaborative Fund managed by the MassTech Collaborative.
\end{acks}

\bibliographystyle{ACM-Reference-Format}
\bibliography{ref}

\appendix
\section{Speaker-Aware Animation network}

The attention network follows the encoder block structure in \citet{vaswani2017attention}. It is composed of a stack of $N=2$ identical layers. Each layer has (a) a multi-head self-attention mechanism with $N_{head}=2$ heads and dimensionality $d_{model}=32$ and (b) a fully connected feed-forward layer whose size is the same to the input size. We also use the embedding layer
(a one-layer MLP with hidden size 32) and the position encoder layer as mentioned in \citet{vaswani2017attention}. 
The discriminator network has a similar network architecture to the attention network. The difference is that the discriminator also includes a three-layer MLP network which has 512, 256, 1 hidden size respectively.

\begin{table}[tb]
\caption{Generator architecture for synthesizing natural face images. \label{tab:stage1}}
\vspace{-3mm}
\centering
\begin{tabular}{c|c}
\toprule
\hline
&Landmark Representation  $\bY_t$ \\ 
$256\times256$&Input Image  $\bQ$\\ 
\hline
$128\times128$ & ResBlock down $(3+3)\rightarrow 64$ \\
\hline
$64\times64$ & ResBlock down $64\rightarrow 128$ \\
\hline
$32\times32$ & ResBlock down $128\rightarrow 256$ \\
\hline
$16\times16$ & ResBlock down $256\rightarrow 512$ \\
\hline
$8\times8$ & ResBlock down $512\rightarrow 512$ \\
\hline
$4\times4$ & ResBlock down $512\rightarrow 512$ \\
\hline
$8\times8$ & ResBlock up $512 \rightarrow 512$ \\
\hline
$16\times16$ & Skip + ResBlock up $(512+512)\rightarrow 512$ \\
\hline
$32\times32$ & Skip + ResBlock up $(512+512)\rightarrow 256$ \\
\hline
$64\times64$ & Skip + ResBlock up $(256+256)\rightarrow 128$ \\
\hline
$128\times128$ & Skip + ResBlock up $(128+128)\rightarrow 64$ \\
\hline
$256\times256$ & Skip + ResBlock up $(64+64)\rightarrow 3$ \\
\hline
$256\times256$ & Tanh \\
\hline
\bottomrule
\end{tabular}
\vspace{-2mm}
\end{table}

\section{Image-to-image translation network}
The layers of the network architecture used to  generate natural human face images are listed in Table \ref{tab:stage1}. In this table, the left column indicates the spatial resolution of the feature map output. 
    \textbf{ResBlock down} means a 2-strided convolutional layer with $3\times3$ kernel followed by two residual blocks,
    \textbf{ResBlock up} means a nearest-neighbor upsampling with a scale of 2, followed by a $3\times3$ convolutional layer and then two residual blocks,
    and
    \textbf{Skip} means a skip connection concatenating the feature maps of an encoding layer and decoding layer with the same spatial resolution.

\end{document}